%% file: main.tex
\documentclass[10pt,twocolumn,letterpaper]{article}

\usepackage[pagenumbers]{cvpr} %

\definecolor{cvprblue}{rgb}{0.21,0.49,0.74}
\usepackage[pagebackref,breaklinks,colorlinks,allcolors=cvprblue]{hyperref}

\input{preamble}

\withsupptrue

\def\paperID{163} %
\def\confName{3DV\xspace}
\def\confYear{2027\xspace}

\title{Spatial-Spectral Coupling as Attention:\\ Scaling Shape Correspondence to High-Resolution Surfaces}

\author{Bence Demeter$^{1}$\thanks{Equal contribution, ordered alphabetically.} \quad Kilian Ehrenreich$^{1*}$ \quad Amelie Kleber$^{1*}$ \quad Ioana Oprisiu$^{1*}$\\
Philipp Spohn$^{1*}$ \quad Christian Wachinger$^{1,2}$ \quad Fabian Bongratz$^{1,2}$ \quad Lennart Bastian$^{3}$\thanks{Correspondence: \href{mailto:l.bastian@imperial.ac.uk}{\nolinkurl{l.bastian@imperial.ac.uk}}.}\\[0.5em]
{\small $^{1}$Technical University of Munich, Germany \quad
$^{2}$Munich Center for Machine Learning, Germany}\\
{\small $^{3}$Imperial College London, United Kingdom}
}

\begin{document}
\setcounter{footnote}{0} %
\maketitle
\input{sec/0_abstract}

\input{sec/1_intro}
\input{sec/2_related_work}

\input{sec/3_background}
\input{sec/4_method}
\input{sec/5_results}
\input{sec/6_conclusion}

{
    \small
    \bibliographystyle{ieeenat_fullname}
    \bibliography{main}
}

\input{sec/A1_protocol}

\input{sec/A2_implementation}
\input{sec/A3_linearized}
\input{sec/A4_qualitative}

\input{sec/A5_patches}

\end{document}

%% file: preamble.tex
\usepackage{colortbl}
\usepackage{tikz}
\usetikzlibrary{calc}

\definecolor{pendingyellow}{RGB}{255,243,163}

\renewcommand{\paragraph}[1]{{\vspace{0.3mm}\noindent \bf #1}.}

\newif\ifwithsupp
\DeclareRobustCommand{\xdocref}[1]{\ifwithsupp\cref{#1}\else\cref*{#1}\fi}
\DeclareRobustCommand{\Xdocref}[1]{\ifwithsupp\Cref{#1}\else\Cref*{#1}\fi}
\DeclareRobustCommand{\supp}[1]{\xdocref{#1}}
\DeclareRobustCommand{\Supp}[1]{\Xdocref{#1}}
\DeclareRobustCommand{\main}[1]{\xdocref{#1}}

\AtEndPreamble{
  \crefname{figure}{Fig.}{Figs.}
  \Crefname{figure}{Fig.}{Figs.}
  \crefname{table}{Tab.}{Tabs.}
  \Crefname{table}{Tab.}{Tabs.}
  \crefname{section}{Sec.}{Secs.}
  \Crefname{section}{Sec.}{Secs.}
  \crefname{equation}{Eq.}{Eqs.}
  \Crefname{equation}{Eq.}{Eqs.}
  \crefname{theorem}{Thm.}{Thms.}
  \Crefname{theorem}{Thm.}{Thms.}
  \crefname{remark}{Rem.}{Rems.}
  \Crefname{remark}{Rem.}{Rems.}
  \crefname{definition}{Dfn.}{Dfns.}
  \Crefname{definition}{Dfn.}{Dfns.}
  \crefname{algorithm}{Alg.}{Algs.}
  \Crefname{algorithm}{Alg.}{Algs.}
  \crefname{proposition}{Prop.}{Props.}
  \Crefname{proposition}{Prop.}{Props.}
}

%% file: sec/0_abstract.tex
\begin{abstract}
Establishing dense correspondence between high-resolution 3D shapes remains challenging due to the spatial--spectral coupling at the core of many functional-map methods, which grows quadratically with mesh resolution.
We observe that this coupling is equivalent to a scaled dot-product attention formulation, with learned descriptors serving as queries and keys and spectral basis functions as values. 
This connection makes efficient attention implementations directly applicable to shape matching and an effective mechanism for scaling spatial-spectral coupling. 
We show that optimized quadratic attention implementations can scale modern deep functional map pipelines to large mesh sizes, improving both runtime and memory consumption, while sub-quadratic methods introduce errors, particularly for non-isometric shape pairs, where an exact coupling proves most necessary.
These insights motivate \textit{Geo-FlashAttention}, which restricts the exact softmax to geometry-selected surface patches and achieves near-linear empirical scaling while preserving correspondence accuracy of exact implementations.
Across six standard benchmarks and three functional-map pipelines, rigorous analysis shows that efficient couplings preserve correspondence accuracy.
At $60k$ vertices, these insights reduce peak training memory from 106.08 to 5.67\,GiB, with Geo-FlashAttention being $1.5\times$ faster per training step than FlashAttention. 
This enables end-to-end learning for correspondence problems on large meshes such as cortical surfaces, where high resolution is required to preserve disease-relevant intricacies of the surface geometry.
The project page is available \href{https://spatial-spectral-attention.github.io/}{here}.
\vspace{-5.0mm}
\end{abstract}

%% file: sec/1_intro.tex
\section{Introduction}
\label{sec:intro}

\begin{figure}[t]
  \centering
  \includegraphics[width=\linewidth]{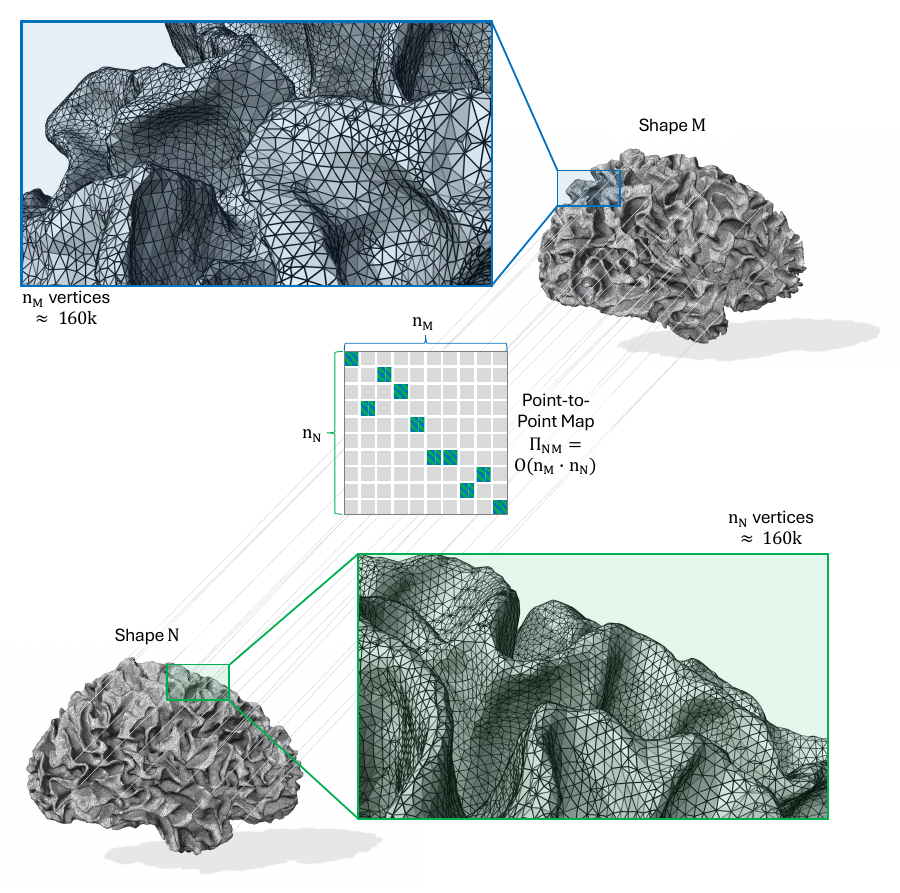}
  \caption{\textbf{Scalability challenges of Deep Spectral Methods.} Spectral matching methods often rely on recovering a point-to-point map $\Pi_{NM}$, either to convert a functional map back to the spatial domain or as an effective regularization.
  Comparing shape vertex features requires $\mathcal{O}(n_M \cdot n_N)$ operations, which is infeasible with current strategies.}
  \label{fig:teaser}
  \vspace{-5.0mm}
\end{figure}

Large-scale 3D meshes with tens of thousands of vertices are often required to represent whole-body avatars or complex anatomical shapes such as cortical surfaces (see \Cref{fig:teaser}).
Establishing fine-grained correspondences between such shapes is a crucial prerequisite for geometry-aware analysis and processing, enabling applications such as avatar animation~\cite{loper2015smpl}, motion transfer~\cite{sumner2004deformation}, and shape editing~\cite{sorkine2007rigid} in computer graphics, as well as intersubject comparison of anatomical variability in biomedical imaging~\cite{Fischl1999fsaverage,bongratz2025v2clong,wang2025beyond,bastian2023s3m}.
In particular, accurate correspondences enable the transformation of \textit{information} across shape instances, allowing local geometric and semantic attributes to be transferred, compared, and statistically analyzed despite substantial variability in shape, pose, or topology~\cite{Amrani2024ssm,ehm2025beyond}.

Many recent advances in non-rigid shape matching build on the functional map framework~\cite{ovsjanikov_functional_2012}, which avoids the combinatorial search over point-to-point maps by representing correspondence as a compact linear operator acting on a truncated Laplace-Beltrami eigenbasis~\cite{reuter2006laplace}.
Learning-based variants~\cite{litany2017deep,donati2020deep} pair this formulation with intrinsic feature extractors~\cite{sharp2022diffusionnet} that learn descriptors in place of hand-crafted spectral signatures~\cite{sun2009concise,aubry2011wave}, employing unsupervised structural or metric-preservation losses~\cite{roufosse2019unsupervised,halimi2019unsupervised}.

A dense point-to-point mapping can then be recovered from a \textit{functional map} in a number of ways, for instance, by finding the nearest spectral embeddings of the source vertices on the target shape~\cite{zhuravlev2026nonrigid} or through refinement procedures~\cite{rodola2015point,vestner2017product,ren2018continuous}.
Such spectral-to-spatial conversion is also an essential algorithmic component of numerous methods, providing structure for optimization in the ambient space for both axiomatic~\cite{melzi_zoomout_2019} and learned methods~\cite{cao2023unsupervised}.

However, current approaches are not directly applicable in the high-resolution regime because the soft assignment manifests as a dense $n \times m$ similarity matrix between the vertices of the two shapes.
Both the forward- and backward-pass require materialization of this matrix, so memory grows quadratically with mesh size: for two shapes with $n = m = 100$k vertices, a single coupling already occupies roughly $40$\,GB in single precision.
This limitation is particularly restrictive in domains where fine geometric detail is essential.
For example, cortical surface analysis routinely employs meshes with more than $100$k vertices per hemisphere~\cite{dale1999cortical,fischl2012freesurfer}, and contemporary digital human avatars often require similarly high resolutions to faithfully capture anatomy and appearance.
Consequently, most existing methods operate on heavily downsampled representations and then transfer predictions to the full-resolution mesh~\cite{maggioli2024rematching}.
However, this introduces approximation errors and potentially discards fine-scale geometric information.

In this work, we revisit the coupling operation at the core of functional-map-based correspondence methods.
We show that the spatial-spectral interaction underlying these models can be reformulated as scaled dot-product attention~\cite{vaswani2017attention}, revealing an unexpected connection between geometric correspondence learning and modern transformer architectures.
Furthermore, we introduce \textbf{Geo-FlashAttention}, a geometry-aware sparse attention mechanism that exploits the local structure of surfaces to achieve near-linear scaling with mesh size, making end-to-end correspondence learning feasible on meshes that were previously beyond reach.
Our main \textbf{contributions} are:

\begin{itemize}
  \item We show that the spatial--spectral coupling $\Pi\Phi_M$ shared by coupled
  functional-map methods is exactly scaled dot-product attention, where the learned
  descriptors act as queries and keys, and the spectral basis as values.
  \item This makes exact efficient-attention algorithms directly applicable. FlashAttention~\cite{dao2022flashattention,dao2024flashattention2} removes the $\mathcal{O}(n_N n_M)$ memory of the coupling, preserving the correspondence up to floating-point rounding.
  \item We show that linearised (kernelised) couplings~\cite{katharopoulos2020transformers,choromanski2021rethinking} are, in contrast, a real approximation.
  They significantly compromise accuracy, which worsens with non-isometry.
  \item We thus introduce \textbf{Geo-FlashAttention}, a block-sparse coupling that evaluates the exact softmax over surface-patch candidates, scales near-linearly, and enables training of coupled methods on high-resolution meshes.
  \item This enables applications such as shape matching for cortical surfaces (over $100$k vertices per hemisphere), where we show that, within the technical advances of our framework, existing functional-map methods can outperform state-of-the-art learned surface-reconstruction methods.
\end{itemize}

%% file: sec/2_related_work.tex
\section{Related Work}
\label{sec:related}
Our approach draws on three distinct bodies of work: non-rigid shape correspondence, efficient attention, and cortical surface registration.
We review each of these subfields, focusing on the functional-map coupling that connects scalable attention to high-resolution correspondence.

\paragraph{Shape correspondence and functional-map coupling} Dense correspondence between non-rigid 3D shapes has been addressed through numerous ways; we outline several trends here, but refer readers to recent surveys for a broader treatment~\cite{zhuravlev2026nonrigid,heuschling2026establishing}. 
Classical approaches include minimum-distortion metric embedding~\cite{bronstein2008numerical}, conformal alignment~\cite{lipman2009mobius,kim2011blended}, and kernel matching; in general, recovering a correspondence under non-isometric deformation while preserving pairwise geometric consistency amounts to an NP-hard quadratic assignment problem~\cite{vestner2017efficient}.
Combinatorial formulations that optimize directly over point-to-point maps~\cite{windheuser2011geometrically,roetzer_spidermatch_2024} provide optimality guarantees but struggle to scale due to this inherent complexity~\cite{amrani2025highres}.
These families trade off accuracy, robustness to non-isometry, supervision requirements, and scalability.

Within this landscape, functional maps~\cite{ovsjanikov_functional_2012} are widely used for dense 3D shape correspondence, especially in unsupervised settings where annotations are difficult to obtain, such as anatomical data~\cite{bastian2023s3m,Amrani2024ssm,bongratz2025v2clong}.
They represent correspondence as a compact linear operator between the eigenbases of two shapes, thereby providing strong structural priors. 
Because of this structure, the framework learns per-vertex descriptors directly from surface features, with supervision~\cite{litany2017deep} or, later, without ground-truth correspondence~\cite{roufosse2019unsupervised}. 

\begin{figure}[t]
  \centering
  \includegraphics[width=\linewidth]{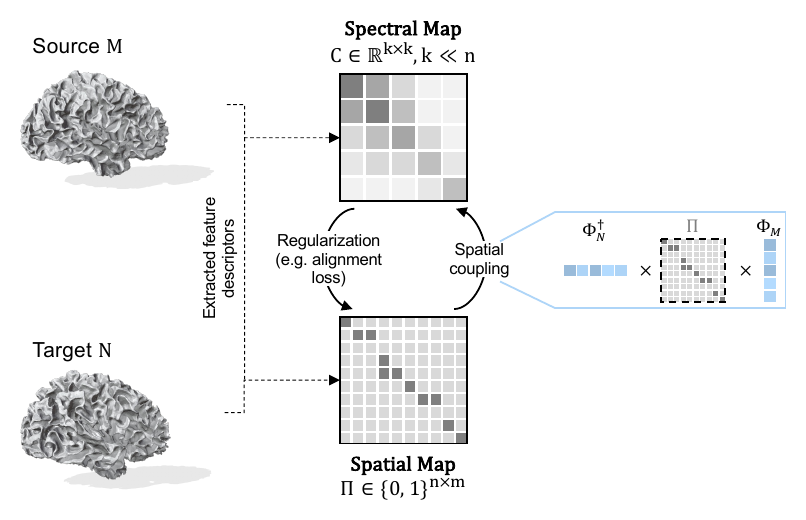}
  \caption{\textbf{Spatial-spectral coupling common in functional map methods \cite{cao2023unsupervised}.} Shared encoders extract feature descriptors of both shapes, which are used to construct functional and point-to-point maps coupled in an alignment regularization.}
  \label{fig:spatial-spectral-coupling}
  \vspace{-6.0mm}
\end{figure}

The standard basis used is the intrinsic Laplace--Beltrami eigenbasis, though hybrid functional maps~\cite{bastian2024hybrid} extend it with an extrinsic elastic component to better handle non-isometric deformations.
The spectral objective alone does not determine a meaningful pointwise map~\cite{sun2023spatially}. Since most applications ultimately require an explicit point-to-point correspondence, high accuracy relies on coupling the operator back to the spatial domain \cite{cao2023unsupervised}.
This coupling is realised either through a soft pointwise map $\Pi = \operatorname{softmax}(F_N F_M^\top / \tau)$ in learned pipelines~\cite{cao2023unsupervised} (\cref{fig:spatial-spectral-coupling}), or through iterative refinement such as ZoomOut~\cite{melzi_zoomout_2019, magnet_memory-scalable_2024}.
The resulting map $\Pi \in \mathbb{R}^{n_N \times n_M}$ is dense, and its $O(n_N n_M)$ memory limits coupled methods to low-resolution meshes. 
\citet{magnet_scalable_2023} scale the classical pipeline to dense meshes by approximating the functional spaces and the map itself through subsampling.
For learned pipelines, \cite{magnet_memory-scalable_2024} avoids the approximation altogether: only the projection $\Pi\Phi_M$ is ever needed, and they compute it without materializing $\Pi$, using the symbolic kernel library KeOps~\cite{charlier2021kernel} to evaluate the reduction on the fly.
This removes the memory bottleneck but remains computationally expensive: every one of the $n_N n_M$ pairwise terms is still evaluated, so the projection remains quadratic in time.

\paragraph{Efficient attention}
Scaled dot-product attention~\cite{vaswani2017attention} is a key building block of modern neural architectures, most apparent in large language models, which popularised it.
Its cost grows quadratically with sequence length, and a large literature reduces this cost along three lines. 
Exact methods keep the softmax but never materialize the full attention matrix, computing the softmax-weighted reduction with hardware-aware tiling (e.g., FlashAttention~\cite{dao2022flashattention, dao2024flashattention2}). 
Approximate methods instead replace the softmax with a kernel feature map, reassociating the product into linear time and memory~\cite{katharopoulos2020transformers, choromanski2021rethinking}. 
A third line restricts each query to a subset of keys through local or block-sparse patterns, keeping the softmax exact over that reduced set while avoiding the quadratic cost of attending to all keys~\cite{beltagy2020longformer, zaheer2020bigbird}. 
However, these partitions are defined by sequence structure rather than by the geometry of the domain.
Transformers have also been used for non-rigid point-cloud registration~\cite{trappolini_shape_2021} and shape correspondence with positional encoding~\cite{raganato_attention_2023}.
\citet{riva_localized_2024} replace some learned self-attention weights with localized Gaussians to improve training efficiency and correspondence quality.
Attention also appears in functional-map methods, applied over descriptor features for partial shapes~\cite{attaiki2021dpfm}, or to combine multiple spectral resolutions~\cite{li2022multiresolution}.
These approaches use attention within correspondence architectures. We focus on the existing soft spatial--spectral projection and evaluate efficient attention kernels for this operation.
Geo-FlashAttention selects geodesic patch support for the descriptor-based softmax, while the Gaussian weights of \citet{riva_localized_2024} define self-attention within a Transformer.

\paragraph{Registration of cortical surfaces}
Establishing correspondences between cortical surfaces is a prerequisite for comparing brain anatomy across subjects and for the population-level analysis of biomarkers such as cortical thickness, which underpins the study of neurodegenerative diseases like Alzheimer's~\cite{Dickerson2008alzheimers}.
Existing methods fall into three broad families.
Spherical registration methods~\cite{Yeo2010demons,Robinson2014msm} inflate each hemisphere to a sphere and align surfaces in this canonical domain; this is the strategy adopted by the widely used FreeSurfer pipeline~\cite{fischl2012freesurfer}.
While robust, the inflation step is computationally intensive and inevitably introduces metric distortions that propagate into the resulting correspondences.
More recently, deformation-based methods~\cite{bongratz2024neural} estimate correspondences by directly deforming surfaces in the ambient 3D space.
Spectral approaches~\cite{lombaert2011fast} instead match surfaces in a low-dimensional embedding derived from the Laplace--Beltrami operator, avoiding the spherical parameterization altogether, but modern feature correspondence methods cannot easily scale to the high resolution demanded by intricate cortical surfaces.

%% file: sec/3_background.tex
\section{Background \& Motivation}
\label{sec:background}

\subsection{Functional Maps and Spectral Constraints}
\label{sec:fmaps}

\begin{figure*}[t]
\vspace{-2.0mm}
\centering
\includegraphics[width=\textwidth]{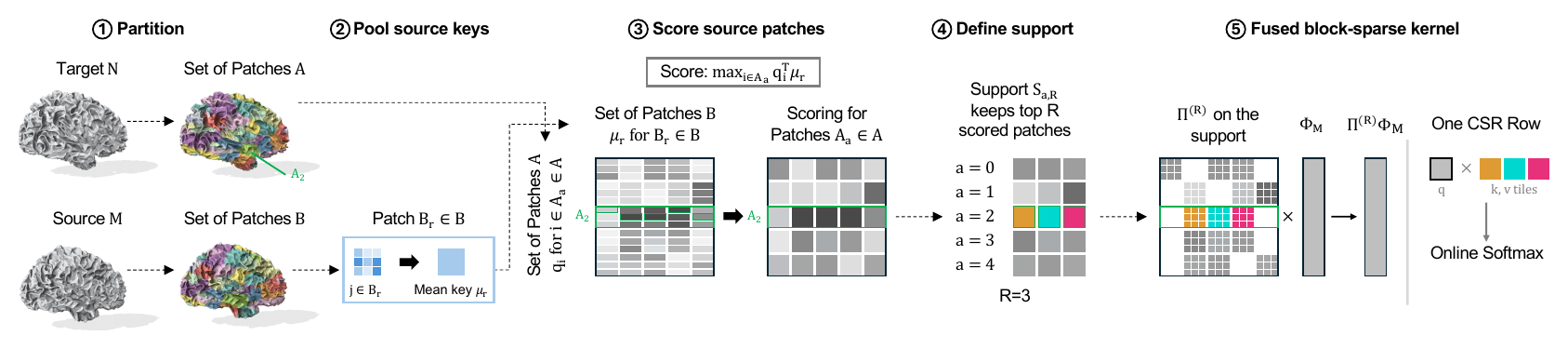}
\caption{
\textbf{Overview of Geo-FlashAttention.}
\textbf{(1)} Target $\mathcal{N}$ and source $\mathcal{M}$ are partitioned into geometrically derived patches $\mathcal{A}$ and $\mathcal{B}$ based on vertex positions.
\textbf{(2)} Each source patch $B_r$ is summarised by its mean key $\mu_r$.
\textbf{(3)} Target queries are scored against the pooled keys and reduced per patch pair by
$s_{ar}=\max_{i\in A_a} q_i^\top \mu_r$.
\textbf{(4)} The $R$ best-scoring source patches per target patch form the support
$\mathcal{S}_{a,R}$, so every row keeps exactly $R$ patches.
\textbf{(5)} A fused block-sparse kernel evaluates the exact softmax without materialising $\widetilde\Pi^{(R)}_{NM}$.
}
\label{fig:geoflash-pipeline}
\vspace{-5.0mm}
\end{figure*}

Functional maps provide a compact spectral representation for dense correspondences between 3D shapes.
Given a source shape $\mathcal{M}$ with $n_M$ vertices and a target shape $\mathcal{N}$ with $n_N$ vertices, let $\Phi_M \in \mathbb{R}^{n_M \times k}$ and $\Phi_N \in \mathbb{R}^{n_N \times k}$ contain the first $k$ Laplace--Beltrami eigenfunctions of the respective shapes.
A functional map $C_{MN} \in \mathbb{R}^{k \times k}$ represents a linear correspondence between functions expressed in these spectral bases: for a function $f_M \approx \Phi_M a$ on $\mathcal{M}$, its corresponding function on $\mathcal{N}$ is represented as
\[
f_N \approx \Phi_N C_{MN} a.
\]
Restricting the representation to the first $k$ eigenfunctions yields a low-dimensional, band-limited approximation of the correspondence, with $k \ll n_M,n_N$.
The functional representation is computationally attractive because the map $C_{MN}$ is of size $k \times k$, and structural priors such as orthogonality, bijectivity, and Laplacian commutativity can be expressed as algebraic penalties in this basis.

\subsection{Spatial-Spectral Coupling}

Optimizing correspondence purely in the spectral domain can be insufficient,
since the spectral training objective does not in general determine a meaningful pointwise correspondence. \citet{sun2023spatially} illustrate this ambiguity with a degenerate solution in which the learned descriptor functions on each shape take the form \(F_i = \Phi_i A_0,\) for some fixed coefficient matrix \(A_0\). Since the spectral descriptor coefficients are \(A_i = \Phi_i^\dagger F_i\), this construction gives \(A_i = A_0\) for every shape. Consequently, the identity functional map \(C_{ij} = I\) satisfies the
descriptor-preservation condition \(C_{ij}A_i = A_j\) for every pair. 
Moreover, the identity map trivially satisfies orthogonality and bijectivity constraints. 
Thus, even a functional map that is optimal under standard spectral objectives need not yield a meaningful point-to-point correspondence.

This ambiguity motivates coupling the spectral representation back to a spatial correspondence. A discrete pointwise map \(\Pi_{NM} \in \mathbb{R}^{n_N \times n_M}\) induces a functional map through the spatial-to-spectral projection
\begin{equation}
\widetilde{C}_{MN} = \Phi_N^\dagger \Pi_{NM}\Phi_M,
\label{eq:spatial-spectral-projection}
\end{equation}
where \(\dagger\) denotes the mass-weighted left inverse. The set of functional maps arising from such pointwise correspondences was formalized by \citet{ren2021discrete} as the space of \emph{proper} functional maps. Enforcing consistency between a spectrally estimated functional map and one induced from a spatial correspondence provides a simple yet effective way to resolve ambiguities left by purely spectral objectives, and forms the basis of several subsequent functional-map methods \cite{cao2023unsupervised,bastian2024hybrid,magnet_memory-scalable_2024,sun2023spatially,attaiki2023understanding}.

For instance, the ULRSSM framework \cite{cao2023unsupervised} enforces this consistency through an alignment loss.
From learned per-vertex descriptors \(F_M \in \mathbb{R}^{n_M \times d}\) and \(F_N \in \mathbb{R}^{n_N \times d}\), it constructs a soft pointwise correspondence
\[
\Pi_{NM} = \operatorname{Softmax}\left(\frac{F_NF_M^\top}{\tau}\right).
\]
Substituting this assignment into Eq.~\eqref{eq:spatial-spectral-projection} gives the induced functional map, which is compared with the map obtained from the spectral solver branch.
Their disagreement is penalized during training, encouraging the spectral solution to remain consistent with a spatially realized correspondence.

Similarly, the ZoomOut algorithm \cite{melzi_zoomout_2019} relies on the same spatial-to-spectral coupling, implementing it through iterative refinement. %
Starting from a functional map at spectral resolution \(k\), ZoomOut recovers a pointwise map and projects it back into a higher-dimensional basis containing the first \(k + \Delta k\) eigenfunctions:

\(C^{(k+\Delta k)} = \left(\Phi_N^{(k+\Delta k)}\right)^\dagger \Pi^{(k)}\Phi_M^{(k+\Delta k)}.\)

Repeating this process ensures that the refined spectral maps are induced by explicit pointwise correspondences while progressively increasing the spectral resolution. 
Recent differentiable variants of ZoomOut \cite{magnet_memory-scalable_2024} replace hard nearest-neighbor assignments with soft correspondences, making this spatial-spectral conversion %
differentiable and thus compatible with end-to-end training.

\subsection{Scalability of Spatial--Spectral Coupling}

The shared computational primitive in both learned functional-map pipelines and refinement methods is the pointwise-to-functional projection 
\(\Phi_N^\dagger \Pi \Phi_M.\)

Its dominant cost is the intermediate product \(\Pi\Phi_M\), where \(\Pi \in \mathbb{R}^{n_N \times n_M}\) is a dense soft correspondence and \(\Phi_M \in \mathbb{R}^{n_M \times k}\) is a truncated spectral basis. Although \(\Pi\) contains \(n_N n_M\) pairwise correspondence weights, only the product \(\Pi\Phi_M\) is required to construct the induced functional map. The dense correspondence is therefore only an intermediate representation.

A naïve implementation nevertheless materializes \(\Pi\), incurring \(O(n_N n_M)\) memory and limiting applicability to high-resolution meshes. \citet{magnet_memory-scalable_2024} observed that this intermediate matrix can be avoided by computing the product \(\Pi\Phi_M\) directly, without explicitly constructing \(\Pi\).
This removes the need to store the dense feature assignment while preserving the resulting point map.

Building on this observation, we show that for standard soft correspondences, this reduction can be written as scaled dot-product attention, up to the choice of scale. 
This allows the spatial-to-spectral projection to be evaluated with highly optimized exact attention kernels, connecting spatial-spectral coupling to efficient attention methods developed for scaling transformers.

%% file: sec/4_method.tex
\section{Spatial--Spectral Coupling as Attention}
\label{sec:method}

The coupling in Eq.~\eqref{eq:spatial-spectral-projection} requires only the
product $\Pi_{NM}\Phi_M$. For the descriptor-based soft assignment, this is
$$
\Pi_{NM} \Phi_M = \operatorname{Softmax}\left(\frac{F_N F_M^\top}{\tau}\right) \Phi_M.
$$
We observe that this formulation can be written as Scaled Dot-Product Attention (SDPA), with scale $1/\tau$ in place of the usual $1/\sqrt{d}$, under the identification
$$
Q = F_N, \qquad K = F_M, \qquad V = \Phi_M.
$$
The learned shape descriptors ($Q, K$) determine the correspondence weights (the attention matrix), while the values ($V$) are the spectral basis functions being transported through that correspondence.

This reformulation makes optimized attention variants directly applicable to spatial-spectral shape matching. 
In particular, we utilize FlashAttention \citep{dao2022flashattention,dao2024flashattention2}, which computes the exact softmax-weighted reduction using hardware-aware tiling, avoiding the explicit materialization of the $n_N \times n_M$ attention matrix. 
This reduces the memory complexity of the coupling layer from $\mathcal{O}(n_N n_M)$ to $\mathcal{O}(n_N + n_M)$, enabling training on high-resolution meshes.

\begin{figure}[t]
  \centering
  \includegraphics[width=\linewidth]{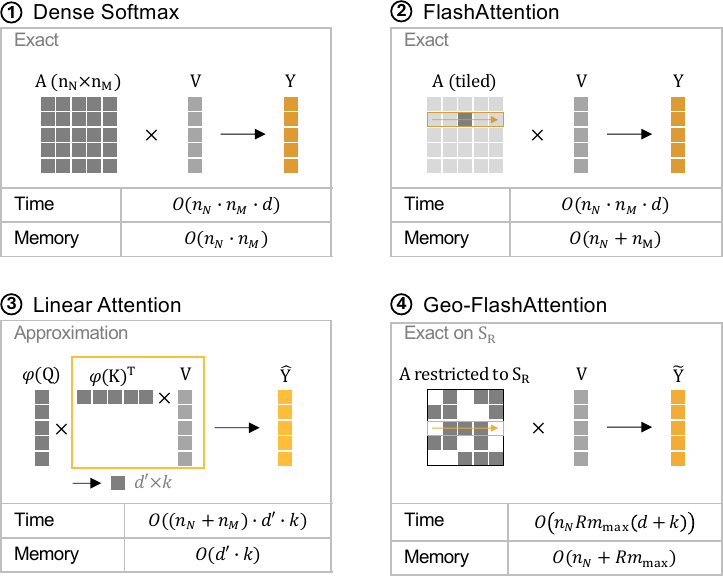}
  \caption{\textbf{Comparing attention computations.} 
  \textbf{(1)} Dense softmax forms the full $n_N\times n_M$ correspondence matrix, while FlashAttention
  \textbf{(2)} tiles $A$ through on-chip memory.
  \textbf{(3)} Linear attention replaces the softmax kernel by a feature map $\varphi$ approximating the product.
  \textbf{(4)} Geo-FlashAttention evaluates exact softmax attention on the induced vertex support by $\mathcal{S}$.
    }
  \label{fig:attention-mechanisms}
  \vspace{-5.0mm}
\end{figure}

\subsection{Approximate Spatial--Spectral Projection}

\paragraph{Linear Attention}
FlashAttention removes the quadratic memory requirement, but exact attention still has quadratic time complexity. 
Linear attention \citep{katharopoulos2020transformers} instead replaces the exponential kernel by a non-negative feature map $\phi:\mathbb R^d\rightarrow\mathbb R^r$ reassociating the product into linear time and memory:
\begin{equation}
 \operatorname{LinAttn}(Q,K,V)_i
 =\frac{\phi(q_i)^\top\sum_{j=1}^{n_M}\phi(k_j)v_j^\top}
 {\phi(q_i)^\top\sum_{j=1}^{n_M}\phi(k_j)}.
    \label{eq:kernel_matmul_assoc}
\end{equation}
The standard choice $\phi(x)=\operatorname{ELU}(x)+1$ is far too diffuse to
approximate the sharply peaked coupling in
Eq.~\eqref{eq:spatial-spectral-projection}, so we sharpen it with a
temperature-matched polynomial factor,
$\phi(x)_\ell=\bigl(\operatorname{ELU}(x_\ell)+1\bigr)\bigl(1+|x_\ell|/\tau\bigr)^{p}$
with $r{=}d$, and additionally evaluate Performer's random-feature softmax
estimator~\citep{choromanski2021rethinking}.
Regardless of the map, the factorized kernel $\phi(Q)\phi(K)^\top$ has rank at
most $r$.
We observe empirically that accurate correspondence estimation requires the soft vertex assignment to
be near-full rank (\supp{app:linear}).
Linearized couplings are therefore lossy at practical
feature dimensions, motivating a sparse alternative.

\subsection{Geo-FlashAttention}
\label{sec:geo-flash}

Let $x_i^N\in\mathbb R^3$ denote the position of target vertex $i$ on shape $\mathcal{N}$, with
normalized query $q_i$ given by the corresponding row of $Q=F_N$. 
Similarly, $x_j^M$ and $k_j$ denote a source vertex position and its normalized key from
$K=F_M$. 
Using geodesic distances on the mesh, we independently partition the target vertex-index set $\mathcal I_N=\{1,\ldots,n_N\}$ and source
vertex-index set $\mathcal I_M=\{1,\ldots,n_M\}$ into patch collections
$\mathcal A$ and $\mathcal B$, respectively (\cref{fig:geoflash-pipeline}),
\begin{equation}
 \mathcal I_N=\bigsqcup_{A_a\in\mathcal A}A_a,
 \qquad
 \mathcal I_M=\bigsqcup_{B_r\in\mathcal B}B_r.
 \label{eq:geo-partitions}
\end{equation}

\begin{table*}[t]
\centering
\caption{
\textbf{Correspondence accuracy} of feature couplings, as mean geodesic error ($\times 100$, $\downarrow$) over three seeds $\pm$ standard deviation.
Within each method block, only the computation of $\Pi\Phi$ changes. Learned-pipeline rows differ due to coupling and optimization convergence. We run Bonferroni-corrected simultaneous $95\%$ Welch confidence intervals for mean differences from dense runs, for which no statistically significant differences are detected (they all include zero). Our proposed couplings are in bold. %
\vspace{-1.0mm}
}
\label{tab:main}

\setlength{\tabcolsep}{4pt}
\resizebox{\textwidth}{!}{%
\begin{tabular}{ll cccccc}
\toprule
\textbf{Method} & \textbf{Coupling}
& \textbf{FAUST} & \textbf{SCAPE} & \textbf{SHREC'19}
& \textbf{SMAL} & \textbf{DT4D-intra} & \textbf{DT4D-inter}
\\
\midrule

\multicolumn{8}{l}{\emph{Iterative refinement}}
\\

ZoomOut \citep{melzi_zoomout_2019, magnet_memory-scalable_2024}
& Dense
& $1.79 \pm 0.00$ & $4.33 \pm 1.38$ & $6.17 \pm 0.13$
& $23.48 \pm 3.72$ & $1.94 \pm 0.04$ & $5.17 \pm 0.18$
\\

& \textbf{FlashAttention}
& $1.79 \pm 0.00$ & $4.18 \pm 1.53$ & $5.70 \pm 0.67$
& $25.51 \pm 2.81$ & $1.87 \pm 0.15$ & $5.13 \pm 0.15$
\\

& \textbf{Geo-FlashAttention}
& $1.80 \pm 0.01$ & $4.48 \pm 2.11$ & $5.71 \pm 0.24$
& $24.29 \pm 1.73$ & $2.03 \pm 0.25$ & $6.48 \pm 2.18$
\\

\midrule

\multicolumn{8}{l}{\emph{Learned pipelines}}
\\

ULRSSM \citep{cao2023unsupervised}
& Dense
& $1.56 \pm 0.04$ & $1.87 \pm 0.02$ & $4.45 \pm 0.98$
& $4.14 \pm 0.35$ & $0.96 \pm 0.05$ & $4.12 \pm 0.33$
\\

& \textbf{FlashAttention}
& $1.58 \pm 0.01$ & $1.88 \pm 0.02$ & $4.29 \pm 0.29$
& $4.01 \pm 0.26$ & $1.00 \pm 0.08$ & $4.01 \pm 0.49$
\\

& \textbf{Geo-FlashAttention}
& $1.57 \pm 0.04$ & $1.87 \pm 0.01$ & $4.65 \pm 0.58$
& $4.08 \pm 0.33$ & $0.93 \pm 0.02$ & $3.78 \pm 0.10$
\\

\cmidrule(lr){1-8}

HybridFMaps \citep{bastian2024hybrid}
& Dense
& $1.36 \pm 0.01$ & $1.79 \pm 0.03$ & $3.94 \pm 0.62$
& $3.68 \pm 0.11$ & $0.96 \pm 0.03$ & $3.49 \pm 0.05$
\\

& \textbf{FlashAttention}
& $1.37 \pm 0.01$ & $1.78 \pm 0.02$ & $3.57 \pm 0.41$
& $3.62 \pm 0.20$ & $1.00 \pm 0.03$ & $3.47 \pm 0.16$
\\

& \textbf{Geo-FlashAttention}
& $1.37 \pm 0.00$ & $1.75 \pm 0.00$
& $4.20 \pm 0.83$ & $3.41 \pm 0.27$
& $1.03 \pm 0.02$ & $3.48 \pm 0.03$
\\

\bottomrule
\end{tabular}%
}
\vspace{-5.0mm}
\end{table*}

We use geodesic farthest-point-sampled centers, followed by geodesic nearest-center assignment, and add centers to split oversized patches.
For each source patch $B_r$, define its mean key
\begin{equation}
 \mu_r=\frac{1}{|B_r|}\sum_{j\in B_r}k_j.
 \label{eq:geo-mean-key}
\end{equation}
A target patch $A_a$ scores source patch $B_r$ by
\begin{equation}
 s_{ar}=\max_{i\in A_a}q_i^\top\mu_r.
 \label{eq:geo-patch-score}
\end{equation}
Let $\mathcal S_{a,R}$ contain the indices of the $R$ largest scores. For
$i\in A_a$, define $\chi_{ij}^{(R)}=1$ if
$j\in\bigcup_{r\in\mathcal S_{a,R}}B_r$, and $0$ otherwise. 
We define geometric attention as the soft assignment restricted to this support:

\begin{equation}
 \widetilde\Pi_{NM,ij}^{(R)}
 =\frac{\chi_{ij}^{(R)}\exp(q_i^\top k_j/\tau)}
 {\sum_{\ell=1}^{n_M}\chi_{i\ell}^{(R)}
  \exp(q_i^\top k_\ell/\tau)}.
 \label{eq:geo-attention}
\end{equation}

The normalized descriptors score target--source patch pairs.
Based on these scores, the top $R$ source patches for each target patch are selected, defining the weights within that support. 
Once the support is fixed, Eq.~\eqref{eq:geo-attention} is the exact softmax on the selected vertices.
Geo-FlashAttention evaluates the resulting block-sparse product $\widetilde\Pi_{NM}^{(R)}\Phi_M$ directly, replacing the dense product in Eq.~\eqref{eq:spatial-spectral-projection} without materializing the approximate soft correspondence matrix $\widetilde\Pi_{NM}^{(R)}$.

If every source patch contains at most $m_{\max}$ vertices, each target row attends to at most $Rm_{\max}$ source vertices.
The restricted attention stage therefore costs $\mathcal O(n_NRm_{\max}(d+k))$, while the coarse selection costs $\mathcal O(n_N|\mathcal B|d)$.
\Cref{fig:attention-mechanisms} depicts the coupling computations.

%% file: sec/5_results.tex
\section{Results}
\label{sec:results}

\paragraph{Overview}
We implement our framework using the \texttt{DeepShapeMatchingKit} \citep{xie2026deepshapematchingkit}, which provides the core PyTorch codebase for our training and evaluation pipelines.
For consistency, all reported timing and peak-memory measurements are taken on a single, exclusively allocated H200 NVL with $140$\,GB of VRAM.

All coupling variants use a shared training framework and, within each experiment,
receive identical inputs and scales; only the evaluation of $\Pi\Phi$ changes.
The dense and linearized variants use PyTorch tensor operations, KeOps~\cite{charlier2021kernel} uses its
symbolic reduction, the fp32 SDPA and FlashAttention variants use PyTorch's SDPA
interface, and Geo-FlashAttention uses a custom-implemented Triton kernel \cite{tillet2019triton}. 
Precisions and backend-specific settings are detailed in \supp{app:hyperparameters}. 
For runtime and memory comparisons, we warm up each backend, synchronize CUDA and clear memory between runs, and run the backends back-to-back. 

We evaluate on complementary benchmarks spanning near-isometric to strongly non-isometric matching. 
For near-isometric matching, we use the FAUST \citep{bogo_faust_2014} dataset, a standard benchmark containing human shapes in various poses, together with SCAPE \citep{anguelov_scape_2005} and the more challenging SHREC'19 \citep{melzi_shrec19_2019}. 
To assess performance under large non-isometric deformations, we evaluate on the SMAL dataset \cite{zuffi2017small}, which contains highly varied quadruped animals across different species, and on DeformingThings4D \citep{li20214dcomplete} in both its intra- and inter-class settings. 
Finally, we use the released TOPKIDS \citep{laehner2016topkids} meshes (with $\approx 8$--$12$k and $\sim 60$k vertices) to evaluate robustness to topological noise and to demonstrate our method's memory efficiency.

We report the mean and standard deviation over pre-declared random seeds. 
Unless noted otherwise, each configuration uses five seeds.

\subsection{Equivalence on Standard Benchmarks}
\label{sec:accuracy}

\paragraph{Setup}
We first assess whether the dense coupling can be replaced within standard evaluation protocols from the
literature \citep{donati2020deep,cao2023unsupervised,bastian2024hybrid}. 
Each benchmark is trained and evaluated using its established training and validation sets. 
Within each method block of \cref{tab:main} we only vary the point-to-point coupling described in \cref{sec:method}, keeping the model, data, and protocol fixed. 
We report the mean and standard deviation for each setting\footnote{This differs from the best-of-$N$ convention used
in prior work \citep{cao2023unsupervised,bastian2024hybrid}; the reporting protocol is
detailed in \supp{app:evaluation_protocol}.}.

\paragraph{Results}
Across ULRSSM, HFM, and ZoomOut, we observe no systematic difference in correspondence accuracy between the alternative and dense couplings on the tested benchmarks. 
The differences in mean error vary in direction across methods and benchmarks, and are generally small relative to the observed seed variability. 
These results support comparable observed accuracy across model families. 

\subsection{Accuracy--Efficiency Trade-offs}
\label{sec:ablation}

\begin{table*}[t]
\centering
\caption{
\textbf{Cost and accuracy} of ULRSSM \cite{cao2023unsupervised} couplings on the two released TOPKIDS datasets \citep{laehner2016topkids}.
Time and peak memory are for one training step at $\approx 12$k and $\approx 60$k, mean $\pm$ standard deviation over five repeats; geodesic error ($\times 100$) uses $25$ test pairs of the released datasets, mean $\pm$ standard deviation over fixed seeds with the best seed in parentheses (seed counts and protocol: \supp{app:evaluation_protocol}).
$^\ast$Unschedulable: $29.9$ days for one $100$-epoch run.
$^\dagger$Dense training at $60$k exceeds $140$\,GB; $106.08$\,GiB is one iteration.
}
\label{tab:efficiency}

\setlength{\tabcolsep}{3pt}
\resizebox{\textwidth}{!}{%
\begin{tabular}{l ccc ccc}
\toprule
&
\multicolumn{3}{c}{\textbf{TOPKIDS} ($\approx 8$--$12$k vertices)}
&
\multicolumn{3}{c}{\textbf{TOPKIDS-HD} ($\approx 60$k vertices)}
\\
\cmidrule(lr){2-4}
\cmidrule(lr){5-7}

\textbf{Coupling}
& \textbf{Time} (ms) $\downarrow$ & \textbf{Peak mem.} (GiB) $\downarrow$ & \textbf{Geo. err.} $\downarrow$
& \textbf{Time} (ms) $\downarrow$ & \textbf{Peak mem.} (GiB) $\downarrow$ & \textbf{Geo. err.} $\downarrow$
\\
\midrule

\multicolumn{7}{l}{\emph{Exact couplings}}
\\

~~Dense (fp32)
& $66.0 \pm 7.9$ & $5.22$ & $16.55 \pm 6.84$ ($5.92$)
& $710.6 \pm 105.6$ & $106.08$ & OOM$^\dagger$
\\

~~KeOps (fp32) \citep{magnet_memory-scalable_2024}
& $1943.2 \pm 7.0$ & $1.23$ & $-$
& $38285.9 \pm 115.5$ & $5.67$ & Unsched.$^\ast$
\\

~~SDPA (fp32)
& $78.9 \pm 5.2$ & $1.23$ & $19.35 \pm 7.41$ ($8.43$)
& $861.5 \pm 5.0$ & $5.67$ & $11.05 \pm 1.69$ ($8.83$)
\\

~~\textbf{FlashAttention} (bf16)
& $46.4 \pm 0.3$ & $1.23$ & $15.65 \pm 6.52$ ($9.08$)
& $207.2 \pm 5.0$ & $5.67$ & $11.38 \pm 3.11$ ($7.02$)
\\

\midrule

\multicolumn{7}{l}{\emph{Approximate couplings}}
\\

~~Performer
& $44.4 \pm 1.8$ & $1.23$ & $37.06 \pm 1.68$ ($34.89$)
& $109.7 \pm 4.8$ & $5.67$ & $37.73 \pm 1.24$ ($36.39$)
\\

~~Linear attention (ELU)
& $\mathbf{43.5} \pm 1.8$ & $1.23$ & $35.10 \pm 1.87$ ($32.70$)
& $\mathbf{108.4} \pm 4.9$ & $5.67$ & $30.60 \pm 0.91$ ($29.31$)
\\

~~\textbf{Geo-FlashAttention} (ours)
& $52.3 \pm 2.0$ & $1.23$ & $13.30 \pm 7.51$ ($7.25$)
& $136.2 \pm 4.6$ & $5.67$ & $8.79 \pm 1.28$ ($6.97$)
\\

\bottomrule
\end{tabular}%
}
\vspace{-5.0mm}
\end{table*}

\paragraph{Setup}
The standard shape-matching benchmarks contain only $5$--$8$k vertices \cite {donati2020deep}. 
At this scale, the coupling only accounts for a small fraction of a training step. 
We therefore assess scalability of ULRSSM \cite{cao2023unsupervised} on the two datasets released with the TOPKIDS benchmark \citep{laehner2016topkids}, using both original ($\approx 8$--$12$k vertices) and the high-resolution ($\approx 60$k) meshes, $26$ unit-area meshes each. 
\Cref{tab:efficiency} reports training-step time, peak memory, and geodesic error. 

\paragraph{Results}
At $12$k, training-step times other than KeOps span $1.8\times$, and every non-dense
method uses $1.23$\,GiB. At $60$k, the corresponding time range increases to
$7.9\times$; dense attention requires $106.08$\,GiB, whereas every non-dense method
uses $5.67$\,GiB. 
FlashAttention is the fastest feasible exact coupling. Because its
kernel is available only in half precision, fp32 SDPA dispatches to the
memory-efficient kernel \citep{rabe_self-attention_2021} and is $4.2\times$ slower
at $60$k. 
The measured KeOps \cite{magnet_memory-scalable_2024} step time would yield approximately one month for a complete run.
This is due to the constant overhead of symbolic reduction, because KeOps materializes a general-purpose fp32 kernel much slower than tuned attention kernels, and forming gradients re-evaluates the unstored reduction, increasing its relative cost in training.

Finally, we fit $y{=}cN^{a}$ over seven decimated resolutions from $5$k to $60$k vertices, yielding isolated empirically estimated scaling exponents of $a{=}1.97$ for dense attention, $1.83$ for FlashAttention, and $0.81$ for Geo-FlashAttention.
The performance crossover is near $20$k vertices, which is far below cortical resolution. 
In \supp{app:scaling} we report the fitting protocol, the full-step exponents, and the complete runtime and memory curves.
On the released low-resolution benchmark, the best fixed seed attains the published ULRSSM value of $9.2$ for every exact coupling ($5.9$--$9.1$) and for Geo-FlashAttention ($7.3$), while both linearized couplings stay far above it ($32.7$ at best).
At both resolutions, Geo-FlashAttention achieves accuracy comparable to exact couplings and substantially lower error than linearized methods.
Variability over seeds is reported in \supp{app:evaluation_protocol}.

\paragraph{Linearized couplings}
FlashAttention and fp32 SDPA are exact kernels and change only the arithmetic, whereas linear attention replaces the softmax kernel with an approximation, and thus changes the correspondence itself.
Measured against the dense reference over five seeds, ELU is significantly worse on four of the six benchmarks and Performer on five, the largest differences occurring on the strongly non-isometric SMAL and DT4D-inter, consistent with the difficulty of representing a sharply peaked coupling with the low-rank feature maps discussed in \cref{sec:method}.
In \Supp{app:linear} we report errors per benchmark and significance levels, and in \supp{app:sharpness} we examine the sharpness that causes this.

\begin{table}[t]
\centering
\caption{
\textbf{Topological noise.} Mean geodesic error ($\times 100$) on the released TOPKIDS
benchmark \citep{laehner2016topkids}, over its $25$ test pairs at their released resolution
($\approx 8$--$12$k) and following existing protocols \cite{cao2023unsupervised,bastian2024hybrid}.
Published rows are quoted as reported.
GeoFA uses Euclidean patches with an error of $4.33 \pm 0.42$ over three seeds.
}
\label{tab:topkids}

\setlength{\tabcolsep}{6pt}
\small
\begin{tabular}{l c}
\toprule
\textbf{Method} & \textbf{Geo. err.} $\downarrow$
\\
\midrule

ZoomOut \citep{melzi_zoomout_2019}                & $33.7$ \\
AttentiveFMaps \citep{li2022multiresolution}           & $23.4$ \\
Deep Shells \citep{eisenberger2020deep}           & $13.7$ \\
Smooth Shells \citep{eisenberger2020smooth}       & $10.8$ \\
ULRSSM \citep{cao2023unsupervised}                & $9.2$  \\
Synchronous Diffusion \citep{cao2024synchronous}  & $5.4$  \\
HFM \citep{bastian2024hybrid,xie2026deepshapematchingkit}                     & $4.6$  \\

\midrule

HFM + \textbf{Geo-FlashAttention} (ours)          & $\mathbf{3.9}$ \\

\bottomrule
\end{tabular}
\vspace{-5.0mm}
\end{table}

\begin{table*}[t]
\vspace{-5.0mm}
\centering
\caption{
\textbf{Label transfer and area distortion on cortical surfaces}, averaged over $90$ validation pairs of the Mindboggle dataset.
Left: manually corrected DKT31 parcellations as preserved by each map.
Right: $D_{\text{area}}(S)$, a map distortion measure. $S$ is the number of geodesic-Voronoi
regions, and $D_{\text{area}}$ growing with $S$.
For Geo-FA, we report the mean over three seeds. All methods are run under identical settings.
$^\ast$FreeSurfer \cite{fischl2012freesurfer} generated the reference labels prior to
manual correction, so it is expected to score strongly on label transfer while confounding the evaluation w.r.t. parcelation labels.
}
\label{tab:cortex_labels}
\label{tab:cortex_area}
\setlength{\tabcolsep}{12pt}
\begin{tabular}{l cc ccc}
\toprule
& \multicolumn{2}{c}{\textbf{Label transfer}}
& \multicolumn{3}{c}{$D_{\text{area}} \downarrow$ \textbf{by regions} $S$}
\\
\cmidrule(lr){2-3}
\cmidrule(lr){4-6}
\textbf{Method}
& \textbf{mIoU} $\uparrow$ & \textbf{Acc.} $\uparrow$
& $\mathbf{10}$ & $\mathbf{40}$ & $\mathbf{200}$
\\
\midrule

FreeSurfer$^\ast$ \citep{fischl2012freesurfer}
& $\mathbf{0.718}$ & $\mathbf{0.851}$
& $0.0601$ & $0.1129$ & $0.1772$
\\

FBM \citep{lombaert2011fast}
& $0.619$ & $0.795$
& $0.0367$ & $0.0783$ & $0.2394$
\\

V2C-Flow \citep{bongratz2024neural}
& $0.568$ & $0.753$
& $0.0823$ & $0.1192$ & $0.2085$
\\

\midrule

ULRSSM \citep{cao2023unsupervised} + \textbf{Geo-FlashAttention}
& $0.633$ & $0.800$
& $\mathbf{0.0184}$ & $\mathbf{0.0325}$ & $\mathbf{0.0706}$
\\

\bottomrule
\end{tabular}
\vspace{-4.0mm}
\end{table*}

\begin{figure*}[t]
    \centering
    \includegraphics[width=\textwidth]{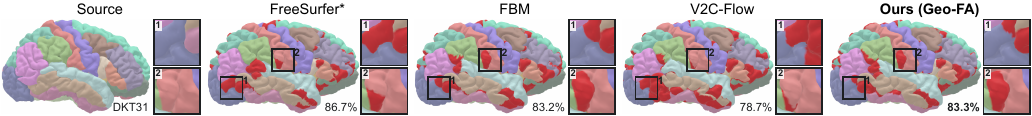}
    \caption{\textbf{Label transfer between two held-out Mindboggle-101 subjects} (lateral pial view).
    Source DKT31 labels are transferred to the target. Disagreements with the target's manually corrected labels are highlighted in red.
    Zoomed regions are depicted in numbered boxes.
    Percentages give vertex accuracy between the pairs.
    $^\ast$Reference labels were initialized with FreeSurfer before manual correction.}
    \label{fig:both}
    \vspace{-4mm}
\end{figure*}

\paragraph{Approximate couplings can act as regularization}
A reduced point-to-point coupling can, in some cases, achieve more than merely approximating its dense counterpart.
Because each query patch distributes its correspondence mass over a small candidate
set rather than over the whole surface (\supp{app:hyperparameters}), implausible
correspondences receive exactly zero weight during training, and the restriction can
act as a regularizer when the input geometry is unreliable. 
We evaluate this on the TOPKIDS benchmark ($\approx 8$--$12$k vertices), where one reference mesh is matched against $25$ scans of children, with their surfaces merged at points of self-contact, thereby probing robustness to topological noise rather than deformation.
\Cref{tab:topkids} compares HFM with Geo-FlashAttention against published results. 
With Euclidean patches, HFM reaches $3.9$, below both reported values for dense-coupled HFM ($5.0$ and $4.6$). \Supp{app:patch_metric} shows how Euclidean patches join surfaces in self-contact.

\subsection{Cortical Surface Matching}
\label{sec:cortex}

Representing the human cortical surface at its native reconstruction resolution requires more than $100$k vertices per hemisphere \citep{dale1999cortical,fischl2012freesurfer}.
At this resolution, dense coupling exceeds practical memory limits (\cref{tab:efficiency}), whereas FlashAttention permits functional-map training on cortical surfaces.

\paragraph{Setup}
Mindboggle-101 \citep{klein2012mindboggle} is the largest set of manually corrected human brain surfaces, providing $101$ subjects with a DKT31 cortical parcellation each. 
We focus on the right-hemisphere white surfaces consisting of $122$--$166$k vertices at native resolution and train on $91 \times 90 = 8{,}190$ mesh pairs derived from $91$ distinct subjects. 
The evaluation is performed on 
the remaining ten validation subjects.
We run ULRSSM \citep{cao2023unsupervised} coupled by Geo-FlashAttention ($R{=}64$), using WKS \cite{aubry2011wave} and anatomical features as input (see \supp{app:cortex}).

\paragraph{Baselines}
FreeSurfer \citep{fischl2012freesurfer} inflates each hemisphere to a sphere and registers it to a folding-based atlas. Fast Brain Matching \citep{lombaert2011fast} (FBM) is the classical spectral correspondence, taken in its geometry-only configuration with no anatomical features. V2C-Flow \citep{bongratz2024neural} deforms the FsAverage template to each subject, so correspondence follows from the shared template topology. \Supp{app:cortex} gives each configuration and how each map is transferred to our meshes.

\paragraph{Metrics}
Cortical surfaces lack ground-truth dense correspondence, so the geodesic error of \cref{tab:main} is unavailable.
We therefore transfer the manually corrected DKT31 parcellation \citep{klein2012mindboggle} ($31$ regions per hemisphere, delineated along folding landmarks) through each map, reporting vertex accuracy and region-averaged IoU (\cref{tab:cortex_labels}).
Because those labels were FreeSurfer-initialized before manual correction, agreement with FreeSurfer's own parcellation is partly built into the reference, so we add a label-free score.
We subdivide $\mathcal{M}$ into $S$ geodesic Voronoi regions, assign each vertex of $\mathcal{N}$ to the region containing its $\Pi_{NM}$-image, and report the mean absolute log-area ratio $D_{\text{area}}(S) = |\mathcal{S}|^{-1} \sum_{s \in \mathcal{S}} \left| \log ( a^{\mathcal{M}}_s / a^{\mathcal{N}}_s ) \right|$ over the regions $\mathcal{S}$ non-empty on both surfaces, where $a^{\mathcal{M}}_s$ and $a^{\mathcal{N}}_s$ are region and preimage areas divided by their native surface's total area.
It is zero for a locally area-preserving map and grows with stretching or compression. \Supp{app:cortex} gives the construction in full.

\paragraph{Results}
Our model attains the lowest area distortion at every scale (\cref{tab:cortex_area}), between $49.8\,\%$ and $60.1\,\%$ below the best baseline.
On label transfer, it ranks second on both scores, ahead of V2C-Flow and FBM but behind FreeSurfer, which is used to generate the reference labels and is therefore favored by that metric. $D_{\text{area}}$ scores only the correspondence map.
\Cref{fig:both} shows the transferred cortical labels and local disagreements with the target reference in two zoomed regions.

%% file: sec/6_conclusion.tex
\section{Conclusion, Limitations \& Future Work}
\label{sec:conclusion}

We identifiy spatial--spectral coupling as an attention operation, allowing efficient kernels to scale functional-map pipelines without sacrificing accuracy. 
Building on this simple yet effective insight, the proposed Geo-FlashAttention demonstrates that geometry-aware sparsification is better suited than linearization, enabling correspondence learning on substantially larger meshes.

\textbf{Limitations and Future Work.} Our reformulation applies only when the correspondence is consumed through a matrix product.
Geo-FlashAttention additionally depends on a fixed candidate budget and lacks approximation guarantees. 
We measure scaling empirically.
Moreover, evaluation on cortical surfaces is limited by the absence of dense ground truth. 
Adaptive candidate selection and stronger evaluation protocols are promising future directions.

\paragraph{Acknowledgements}
The authors are grateful to the TU Munich Data Innovation Lab, which generously hosted this project.
LB was supported by the UK Royal Society through grant NIF/R1/254128.
The authors declare no conflicts of interest.

%% file: sec/A1_protocol.tex
\clearpage
\setcounter{page}{1}
\maketitlesupplementary
\setcounter{figure}{0}
\setcounter{table}{0}
\renewcommand{\thefigure}{S\arabic{figure}}
\renewcommand{\thetable}{S\arabic{table}}

We organize the supplementary materials as follows.
\Cref{app:evaluation_protocol} describes our evaluation protocol for comparison of attention kernels. \Cref{app:hyperparameters}
lists the software, backends, architectures, losses, and learning rate schedules. 
\Cref{app:cortex} provides the Mindboggle-101 evaluation protocol, configurations and the
label-transfer metric behind \main{tab:cortex_labels}. 
\Cref{app:linear} quantifies what the linearized couplings cost in accuracy and coupling rank. 
In \Cref{app:scaling}, we assess scaling characteristics of the different couplings across seven decimated TOPKIDS resolutions. 
In \Cref{app:sharpness}, we analyze whether the fixed softmax temperature of linear attention variants is optimal and whether it can be learned.
\Cref{app:qualitative} shows qualitative transfers across spatial--spectral couplings and cortical correspondence methods.
\Cref{app:patch_metric} assesses the Euclidean patch construction against an improved geodesic patch variant it approximates. 
In \Cref{app:R_budget} we sweep the candidate budget $R$ across benchmarks.
\Cref{app:pck} presents the geodesic error distributions for \main{tab:main,tab:linear}.

\section{Evaluation and Implementation Details}

\subsection{Evaluation Protocol}
\label{app:evaluation_protocol}

\begin{figure}[t]
\centering
\setlength{\tabcolsep}{1pt}

\begin{tabular}{@{}ccc@{}}
{\small Source} & {\small Clean map} & {\small Flipped map} \\
\includegraphics[width=0.315\linewidth]{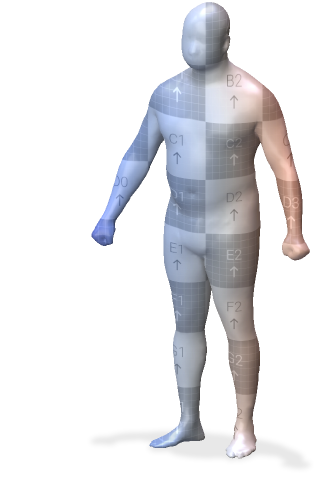} & \includegraphics[width=0.315\linewidth]{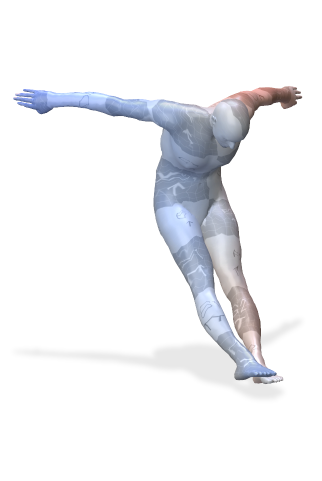} &
\includegraphics[width=0.315\linewidth]{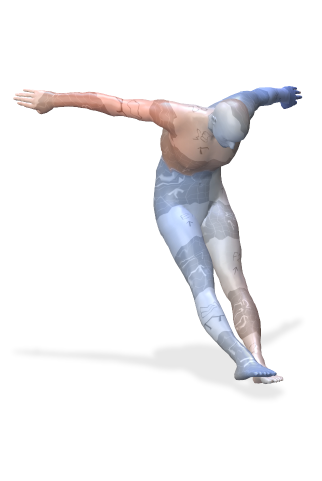} \\
\end{tabular}
\caption{\textbf{Seed-dependent flip on SHREC'19 shape $28$.} The source is coloured
by signed lateral coordinate and pulled through the predicted map. One seed preserves
the sides, while another exchanges them over the upper body.}
\label{fig:shrec19_flip}
\end{figure}

\paragraph{Seeds and statistics}
The deep functional maps (DFM) literature commonly reports the best of $N$ runs.
As we seek to distinguish between computational modules and their effect in various algorithms, we instead report the mean and standard deviation over five pre-declared seeds unless noted otherwise. 
We observe that SGD optimization in the deep variants sometimes converges to discontinuous part-exchanging maps despite comparable training objectives (\cref{fig:shrec19_flip}),
a known failure mode of DFM methods \cite{donati_deep_2022,donati_complex_2022}. 
On SMAL and DT4D-inter this variability is concentrated in pairs containing one bistable shape, \texttt{horse\_05} and \texttt{GoalkeeperScoop}, respectively, and occurs across couplings rather than identifying a particular kernel. 
Excluding these pairs reduces the errors of the learned methods, making their behavior more predictable and enabling easier comparison between runs.
We thus apply this exclusion identically to every row to separate the kernel comparison from the finite-seed flip draw, and report the all-pairs values alongside them in \cref{tab:bistable}. 
SHREC'19 likewise contains visibly bistable shapes, but in the main-table results, we retain all shape pairs as the variance is generally high. 
Separately, one isolated ULRSSM dense seed on SMAL was replaced under a pre-declared, label-free within-arm rule: its closest-peer map disagreement exceeded $1.4\times$ the arm median.

\begin{table}[t]
\centering
\caption{\textbf{All-pairs results.} Mean geodesic error ($\times 100$,
$\downarrow$) including the bistable pairs omitted from the corresponding
SMAL and DT4D-inter cells of \main{tab:main,tab:linear}. The exclusion consistently improves the mean across learned pipelines. It does not improve ZoomOut on SMAL, whose errors are distributed broadly rather than concentrated on the bistable pairs.}
\label{tab:bistable}
\footnotesize
\setlength{\tabcolsep}{5pt}
\begin{tabular}{ll cc}
\toprule
\textbf{Method} & \textbf{Coupling} & \textbf{SMAL} & \textbf{DT4D-inter} \\
\midrule
ZoomOut & Dense              & $23.23 \pm 4.02$ & $8.03 \pm 2.49$ \\
        & FlashAttention     & $25.37 \pm 2.91$ & $9.25 \pm 2.32$ \\
        & Geo-FlashAttention & $24.20 \pm 1.95$ & $11.22 \pm 0.85$ \\
\midrule
ULRSSM  & Dense              & $5.12 \pm 0.70$ & $4.49 \pm 0.29$ \\
        & FlashAttention     & $4.64 \pm 0.69$ & $6.17 \pm 2.26$ \\
        & Geo-FlashAttention & $5.43 \pm 0.57$ & $5.00 \pm 1.96$ \\
        & Linear (ELU)       & $6.82 \pm 0.87$ & $8.19 \pm 1.43$ \\
        & Performer          & $7.54 \pm 1.26$ & $10.94 \pm 0.97$ \\
\midrule
HFM     & Dense              & $4.07 \pm 0.70$ & $3.72 \pm 0.11$ \\
        & FlashAttention     & $3.98 \pm 0.44$ & $3.72 \pm 0.16$ \\
        & Geo-FlashAttention & $3.62 \pm 0.24$ & $3.67 \pm 0.02$ \\
\bottomrule
\end{tabular}
\end{table}

\paragraph{TOPKIDS}
TOPKIDS has no bistable shape, so no pairs are excluded there: \main{tab:efficiency} reports all $25$ test pairs.
On the released benchmarks every coupling uses the same five seeds.

%% file: sec/A2_implementation.tex
\subsection{Implementation Details and Hyperparameters}
\label{app:hyperparameters}

\paragraph{Software and exact backends}
We use PyTorch 2.3.0 with CUDA 12.1. %
KeOps evaluates the same sum-softmax-weight reduction as an fp32 \texttt{LazyTensor}, per the authors' original implementation. 
fp32 SDPA and bf16 FlashAttention utilize PyTorch's scaled dot-product attention interface.

\paragraph{Linearized backends}
Performer uses $256$ orthogonal random features in fp32, with the projection fixed by seed $42$ across runs. 
Linear attention uses the amplified ELU feature map of \main{sec:method} with polynomial amplification $p{=}5$.

\paragraph{Differentiable ZoomOut}
ZoomOut is a standard refinement procedure, but in the differentiable variant the descriptors it refines are learned.
We use ten refinement iterations, growing the spectral dimension from $20$ to $120$ in steps of $10$, with
blur $\sigma{=}10^{-2}$ normalized by the maximum pairwise feature distance.
It uses $128$ eigenfunctions and a DiffusionNet of width $256$ with four blocks, no dropout and a $32$-dimensional output, trained with Adam at $10^{-4}$, $\beta{=}(0.9,0.99)$ and
no weight decay, the learning rate dropping tenfold after the fifth of ten epochs. 
Its loss is the unweighted sum of three terms: the orthogonality defect of the initial $20\times20$ map, that of the refined $120\times120$ map, and the disagreement between the initial map and the leading $20\times20$ block of the refined one, which is
detached. 
Each term is normalized by $100/k^2$ for a given $k$.
We keep ZoomOut's Gaussian coupling $\operatorname{Softmax}(-\lVert q-k\rVert^2/(2\sigma^2))$. 
To run it through the attention backends, we augment the queries and keys so that the negative squared distance is equivalent to a dot product. 
Exact SDPA then evaluates the same Gaussian weights, and Geo-FlashAttention restricts them to its sparse support.

\paragraph{Training}
We use the standard implementations provided by \texttt{DeepShapeMatchingKit} \citep{xie2026deepshapematchingkit}.
ULRSSM and HFM use a DiffusionNet \cite{sharp2022diffusionnet} with width $128$, four blocks, and dropout, which maps $128$-dimensional auto-scaled WKS channels (raw, augmented vertex coordinates on SMAL, as in the reference implementation) to $256$-dimensional descriptors.
Both the WKS descriptors and DiffusionNet's spectral diffusion use the first $128$ Laplace--Beltrami eigenpairs, while the functional map is solved at $200$ eigenfunctions on every benchmark. 
Optimizer settings are as follows: we use Adam at learning rate $10^{-3}$, with the default $\beta{=}(0.9,0.999)$ and no weight decay, cosine-annealed to $10^{-4}$ over the run without warmup, on one pair per step.

\begin{table}[t]
\centering
\small
\setlength{\tabcolsep}{4pt}
\resizebox{\columnwidth}{!}{%
\begin{tabular}{l rrrc rr}
\toprule
& \multicolumn{4}{c}{\textbf{ULRSSM / HFM}} & \multicolumn{2}{c}{\textbf{diff. ZoomOut}}
\\
\cmidrule(lr){2-5}
\cmidrule(lr){6-7}
\textbf{Benchmark} & pairs & epochs & iters & TTA & pairs & iters
\\
\midrule
FAUST     & $6\,400$ & $15$  & $96\,000$ & $0$          & $3\,160$ & $31\,600$ \\
SCAPE     & $2\,601$ & $30$  & $78\,030$ & $0$          & $1\,275$ & $12\,750$ \\
SHREC'19  & $9\,001$ & $10$  & $90\,010$ & $12$         & $4\,435$ & $44\,350$ \\
SMAL      & $841$    & $20$  & $16\,820$ & $15$         & $406$    & $4\,060$  \\
DT4D-H    & $8\,879$ & $10$  & $88\,790$ & $0$\,/\,$15$ & $8\,702$ & $87\,020$ \\
TOPKIDS   & $676$    & $100$ & $67\,600$ & $15$         & ---      & ---       \\
\bottomrule
\end{tabular}%
}
\caption{Training schedule per benchmark, batch size one. SHREC'19 trains on FAUST and
SCAPE; one DT4D training serves the intra- and inter-class evaluations ($0$ and $15$
test-time adaptation (TTA) steps). Differentiable ZoomOut trains ten epochs everywhere,
with no TTA beyond its ten refinement iterations, and is not run on TOPKIDS.}
\label{tab:training_schedule}
\end{table}

\subsection{Cortical Evaluation Details}
\label{app:cortex}

\paragraph{Dataset and surfaces}
Mindboggle-101 pools $101$ T1-weighted MRI scans from several collections, so
acquisition site and protocol vary across the set. 
The white and pial surfaces are FreeSurfer reconstructions \citep{fischl2012freesurfer}, and the DKT31 labels were
produced by FreeSurfer's automated parcellation and then corrected by hand
\citep{klein2012mindboggle}. 
Every method uses the same FreeSurfer reconstructions, except V2C-Flow, which is itself a reconstruction tool.
The ten held-out subjects yield $90$ shape pairs, with each subject in $18$ of them.

\paragraph{Configurations}
For our comparison in \main{tab:cortex_labels} we use ULRSSM \citep{cao2023unsupervised} with $200$ Laplace--Beltrami
eigenfunctions and a DiffusionNet extractor on $134$ anatomical channels (WKS, curvature, FreeSurfer sulcal depth and cortical thickness), coupled by geodesic Geo-FlashAttention ($R{=}64$) at $\tau{=}0.07$, averaged over three seeds.
Each run is scheduled for three epochs ($24\,570$ pairs).
Validation runs every $500$ iterations on the ten held-out subjects, scoring the parcellation Jaccard of the transferred labels.
Early stopping is conducted with a patience of $25$. 
We evaluate these final checkpoints on the same validation subjects.
Among the baselines, FreeSurfer \citep{fischl2012freesurfer} matches vertices on the
registered sphere, selecting the source vertex with the largest barycentric weight.
FBM \citep{lombaert2011fast} uses $K{=}50$ graph-Laplacian eigenvectors and no anatomical features; and V2C-Flow \citep{bongratz2024neural} deforms the $163{,}842$-vertex FsAverage template, and its correspondence is transferred to our meshes by nearest-neighbor lookup.

\paragraph{Label transfer}
The Desikan--Killiany--Tourville protocol assigns each vertex to one of $31$ cortical regions per hemisphere, delineated along folding landmarks \citep{klein2012mindboggle}.
Accuracy is the fraction of vertices whose transferred label matches the manual one.
mIoU intersects the transferred and manual vertex sets for each region and averages them across regions, so small and large regions count equally. 
Because the labels were FreeSurfer-initialized before manual correction, agreement with FreeSurfer's own parcellation is partly built into the reference, which is why we report the label-free $D_{\text{area}}$ beside them.

\paragraph{Area distortion}
The $S$ regions partitioning $\mathcal{M}$ are geodesic Voronoi cells based on farthest-point samples on the input meshes (unrelated to the candidate patches of \main{sec:geo-flash}). 
We partition both surfaces by the lumped vertex mass $M_{vv}$ (\main{sec:fmaps}), so no area is lost at region boundaries, and we average the metric over all non-empty regions on both surfaces. Each region area is divided by its native surface's total area before taking the log ratio.

%% file: sec/A3_linearized.tex
\section{Additional Empirical Results}

\begin{figure*}[t]
\centering
\includegraphics[width=\textwidth]{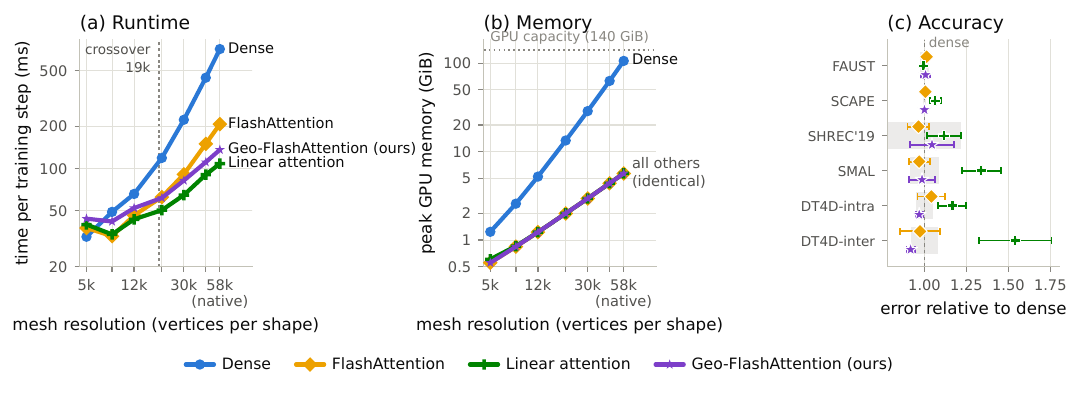}
\caption{Training time and peak memory across seven TOPKIDS resolutions, and accuracy relative to the dense coupling on the standard benchmarks.
In (c), markers are means over five seeds, error bars their standard deviation, and the grey band the spread of the dense reference; the rows repeat the ULRSSM block of \main{tab:main} and the ELU column of \cref{tab:linear} under the same seeds and exclusions (\cref{app:evaluation_protocol}).
Non-dense methods share a single memory curve.
One representative per class of coupling is shown; Performer, KeOps and fp32 SDPA are omitted.}
\label{fig:scaling}
\end{figure*}

\subsection{Cost of Linearized Couplings}
\label{app:linear}

Linear attention replaces the softmax kernel with an approximate computation. 
Unlike the exact kernels of \main{tab:main}, it therefore changes the correspondence computations themselves, and affects the resulting training dynamics. 
We report the performance drop across standard benchmarks in \Cref{tab:linear}.

\begin{table}[t]
\centering
\caption{
\textbf{Cost of linearized couplings.}
Mean geodesic error ($\times 100$, $\downarrow$); unlike \main{tab:main}, these approximate the softmax kernel and change the correspondence.
Markers: two-sample $t$-test against the dense reference over five seeds,
$^{*}p<0.05$, $^{**}p<0.01$, $^{***}p<0.001$.
}
\label{tab:linear}

\setlength{\tabcolsep}{5.4pt}
\resizebox{\columnwidth}{!}{%
\begin{tabular}{l ccc}
\toprule
& \textbf{Dense} & \multicolumn{2}{c}{\textbf{Linear attention}}
\\
\cmidrule(lr){2-2}
\cmidrule(lr){3-4}
\textbf{Benchmark}
& \textbf{(reference)} & \textbf{ELU} & \textbf{Performer}
\\
\midrule

FAUST
& $1.56 \pm 0.04$ & $1.55 \pm 0.01$ & $1.93 \pm 0.13^{***}$
\\
SCAPE
& $1.87 \pm 0.02$ & $1.99 \pm 0.06^{**}$ & $2.47 \pm 0.10^{***}$
\\
SHREC'19
& $4.45 \pm 0.98$ & $4.97 \pm 0.46$ & $5.64 \pm 1.42$
\\
SMAL
& $4.14 \pm 0.35$ & $5.54 \pm 0.48^{***}$ & $6.85 \pm 1.15^{***}$
\\
DT4D-intra
& $0.96 \pm 0.05$ & $1.12 \pm 0.08^{**}$ & $1.29 \pm 0.12^{***}$
\\
DT4D-inter
& $4.12 \pm 0.33$ & $6.35 \pm 0.89^{***}$ & $9.37 \pm 1.06^{***}$
\\

\bottomrule
\end{tabular}%
}
\vspace{-5.0mm}
\end{table}

ELU is similar to the dense coupling on FAUST but significantly worse on four of the six benchmarks.
Performer is notably worse on five. 
Both comparisons are inconclusive on SHREC'19, where seed variance is large. 
The largest gaps are on SMAL and DT4D-inter, the most significantly non-isometric datasets.
This is consistent with the difficulty of representing a sharply peaked softmax with the low-rank feature maps of \main{sec:method}, which we further examine in \cref{app:sharpness}.

\paragraph{Rank of the converged coupling}
In \main{sec:method}, we state that the soft assignment is typically near-full rank.
This is an empirical observation, which we observe in the following settings.
On one converged FlashAttention checkpoint per benchmark and $20$--$40$ test pairs, we replace the exact product $\Pi\Phi$ by each approximation and report the relative Frobenius error of the induced $200\times200$ functional map
(\cref{tab:coupling_rank}). 
Rank is near-full everywhere: even the \emph{optimal} rank-$m$ truncation of $\Pi$ by SVD errs by $0.36$--$0.59$ on every benchmark at $m{=}200$, the spectral dimension, and reaching ${\sim}10^{-2}$ requires $m{=}1024$.

With increasing deviation from isometry, the effective row support grows from $24$ vertices on FAUST to $54$ on SMAL. Over the same range, top-$k$ truncation error increases $5\times$ and spherical-kernel error $2.3\times$, whereas optimal truncation and Nystr\"om show no corresponding degradation. 
We study the associated sharpness in more detail in \cref{app:sharpness}, and \cref{tab:linear} shows the effect on training dynamics in terms of the final geodesic error.

\begin{table}[t]
\centering
\caption{
\textbf{Approximating a converged coupling.}
Each method approximates the converged point-to-point coupling $\Pi$. We assess optimal
rank-$m$ truncation by SVD, Nystr\"om with $m$ landmarks, top-$k$ row truncation, or the spherical
kernel ($p{=}32$). Entries report the relative Frobenius error of the functional map
induced by the approximated coupling, for one converged checkpoint per benchmark.
Support is the effective number of vertices per softmax row. Benchmarks are ordered by
support; best per column in bold.
}
\label{tab:coupling_rank}
\setlength{\tabcolsep}{4pt}
\resizebox{\columnwidth}{!}{%
\begin{tabular}{l c ccccc}
\toprule
& & \multicolumn{2}{c}{\textbf{SVD}} & \textbf{Nystr\"om} & \textbf{top-}$k$ & \textbf{spherical}
\\
\cmidrule(lr){3-4}
\cmidrule(lr){5-5}
\cmidrule(lr){6-6}
\cmidrule(lr){7-7}
\textbf{Benchmark} & \textbf{support}
& $m{=}200$ & $m{=}1024$ & $m{=}1024$ & $k{=}128$ & $p{=}32$
\\
\midrule
FAUST & $24.1$ & $0.449$ & $0.0209$ & $0.149$ & $\mathbf{0.0027}$ & $\mathbf{0.028}$
\\
SCAPE & $24.0$ & $0.459$ & $0.0133$ & $0.160$ & $0.0062$ & $0.038$
\\
DT4D & $36.0$ & $0.587$ & $0.0233$ & $\mathbf{0.126}$ & $0.0113$ & $0.041$
\\
SMAL & $54.4$ & $\mathbf{0.358}$ & $\mathbf{0.0058}$ & $0.157$ & $0.0134$ & $0.064$
\\
\bottomrule
\end{tabular}%
}
\end{table}

\begin{figure*}[!t]
\centering
\setlength{\tabcolsep}{0.8pt}
\renewcommand{\arraystretch}{0.5}

\begin{tabular}{@{}c@{\hspace{2pt}}cccccc@{}}
 & {\small Source} & {\small Dense} & {\small FlashAtt.} & {\small Geo-FlashAtt.}
   & {\small Linear (ELU)} & {\small Performer} \\
\rotatebox{90}{\small\ SMAL} &
  \includegraphics[width=0.156\textwidth]{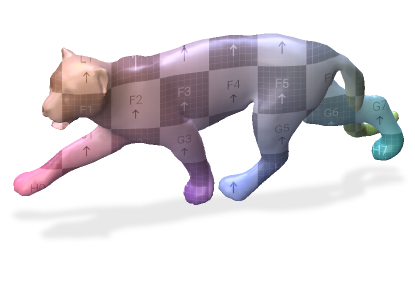} & \includegraphics[width=0.156\textwidth]{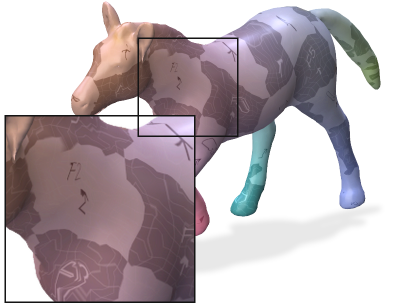} &
  \includegraphics[width=0.156\textwidth]{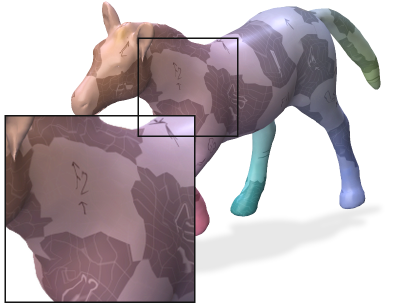} & \includegraphics[width=0.156\textwidth]{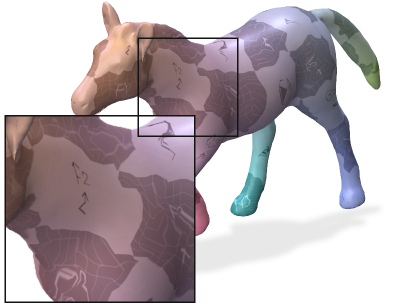} &
  \includegraphics[width=0.156\textwidth]{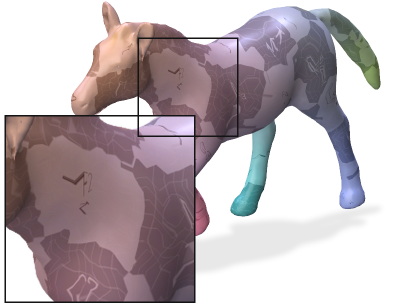} & \includegraphics[width=0.156\textwidth]{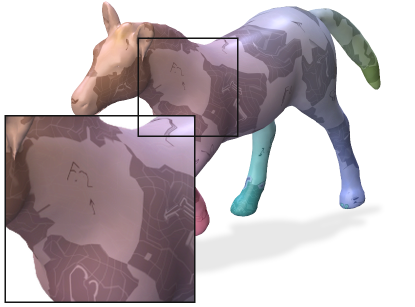} \\
\rotatebox{90}{\small DT4D-inter} &
  \includegraphics[width=0.156\textwidth]{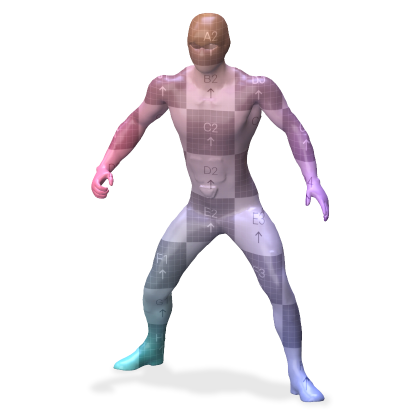} & \includegraphics[width=0.156\textwidth]{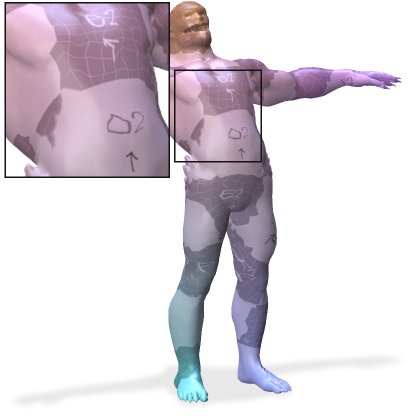} &
  \includegraphics[width=0.156\textwidth]{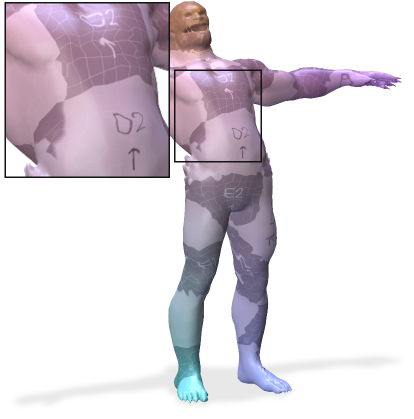} & \includegraphics[width=0.156\textwidth]{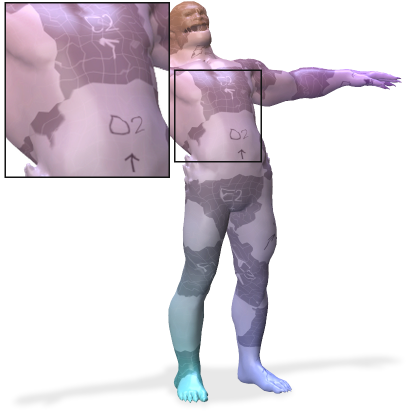} &
  \includegraphics[width=0.156\textwidth]{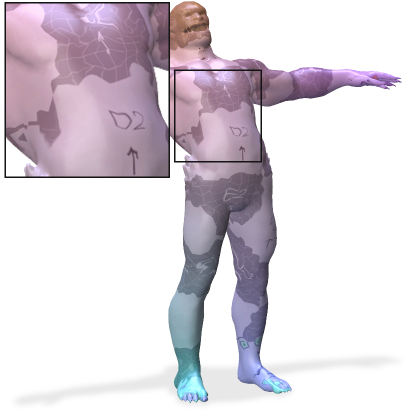} & \includegraphics[width=0.156\textwidth]{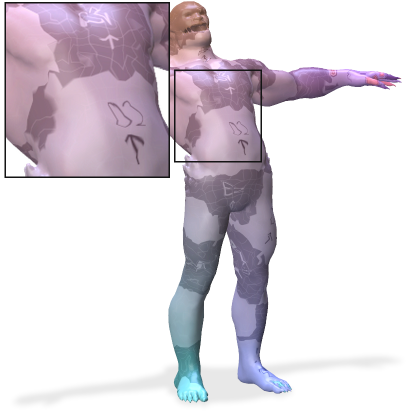} \\
\rotatebox{90}{\small\ FAUST} &
  \includegraphics[width=0.156\textwidth]{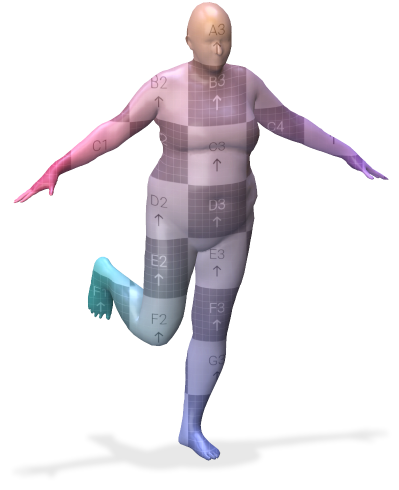} & \includegraphics[width=0.156\textwidth]{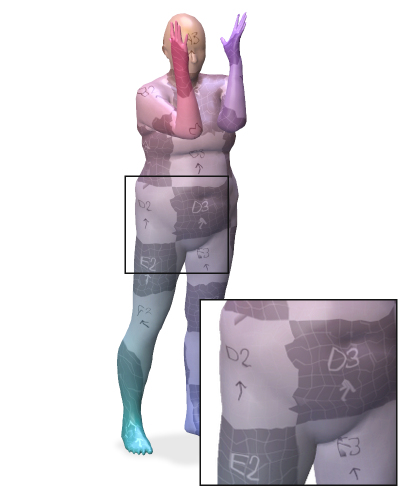} &
  \includegraphics[width=0.156\textwidth]{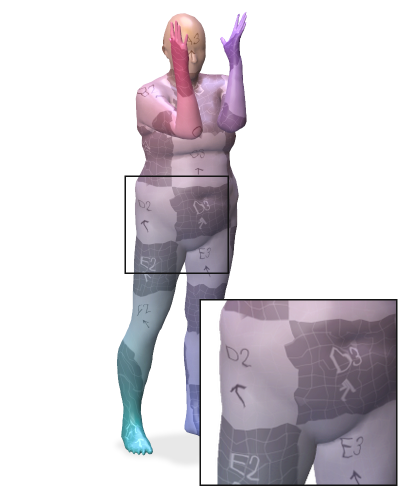} & \includegraphics[width=0.156\textwidth]{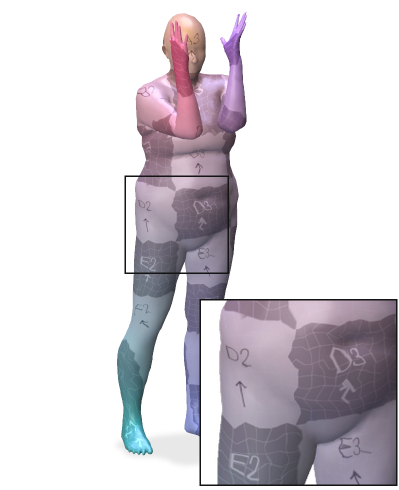} &
  \includegraphics[width=0.156\textwidth]{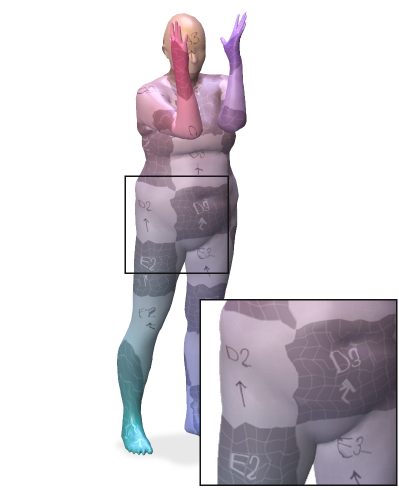} & \includegraphics[width=0.156\textwidth]{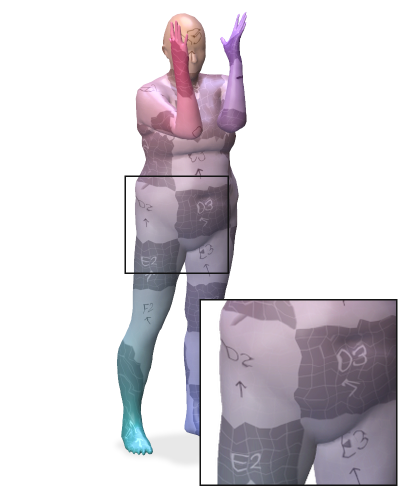} \\
\end{tabular}
\caption{\textbf{Qualitative comparison of spatial--spectral couplings.}
Texture is transferred through maps estimated from the same shape pair and seed.
The exact couplings preserve the same regions. On SMAL and DT4D-inter, Performer
fragments the transferred grid, whereas ELU mainly shifts region boundaries, consistent
with \cref{tab:linear}. All couplings appear similar on FAUST.}
\label{fig:qualitative_couplings}
\end{figure*}

\subsection{Empirical Scaling with Mesh Resolution}
\label{app:scaling}

\paragraph{Setup}
To assess the scaling complexity of Geo-FlashAttention compared to other methods, we decimate the released TOPKIDS-HD scans to six uniform resolutions from $5$k to $45$k vertices, rescaled to unit area.
Together with the native scans ($58$k vertices on average, quoted as $60$k in the main text), this yields seven resolutions. 
At each resolution, we time the coupling and the full training step on one H200~NVL, and fit the exponent $a$ in $y=cN^a$ by least squares on $\log y$ against $\log N$ over the four upper resolutions ($20$k, $30$k, $45$k and native), properly accounting for startup overheads. 
We report scaling coefficients from empirical fits over this decimation range.

\paragraph{Results}
At $5$k the methods differ by only $1.3\times$ per training step, because the feature extractor dominates the computation cost.
At $60$k, we see the coupling dominating. 
The fitted runtime exponents of the isolated coupling are $a{=}1.97$ for dense attention, $1.83$ for FlashAttention, $0.81$ for linear attention and $0.81$ for Geo-FlashAttention.
The full-step exponents are $1.67$, $1.13$, $0.73$ and $0.74$, as feature extraction and spectral operations add
fixed costs. 
Peak training memory scales with $a{=}1.94$ for dense attention and $a{=}0.98$ for every non-dense method, which stay within $0.01$\,GiB of one another at $60$k.
Candidate construction makes Geo-FlashAttention slower at small resolutions, but
it crosses FlashAttention near $20$k vertices. 
At $60$k, its coupling is $3.0\times$, and its training step is $1.5\times$ faster, at the same peak memory. 
\Cref{fig:scaling} shows the full curves.

\begin{figure}[t]
    \centering
    \includegraphics[width=\columnwidth,trim=0 0 180bp 0,clip]{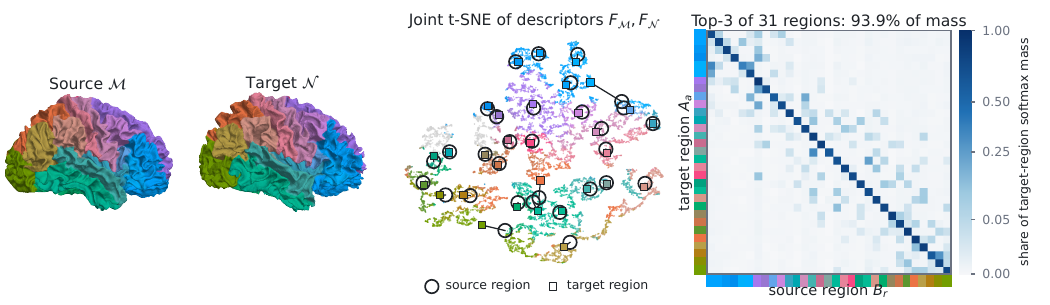}
    \caption{\textbf{Learned cortical descriptors.} One Mindboggle pair, colored by DKT regions, with the joint t-SNE embedding of its descriptors.}
    \label{fig:feature-tsne}
\end{figure}

\begin{figure}[!htb]
    \centering
    \begin{subfigure}[t]{0.30\linewidth}
        \includegraphics[width=\linewidth]{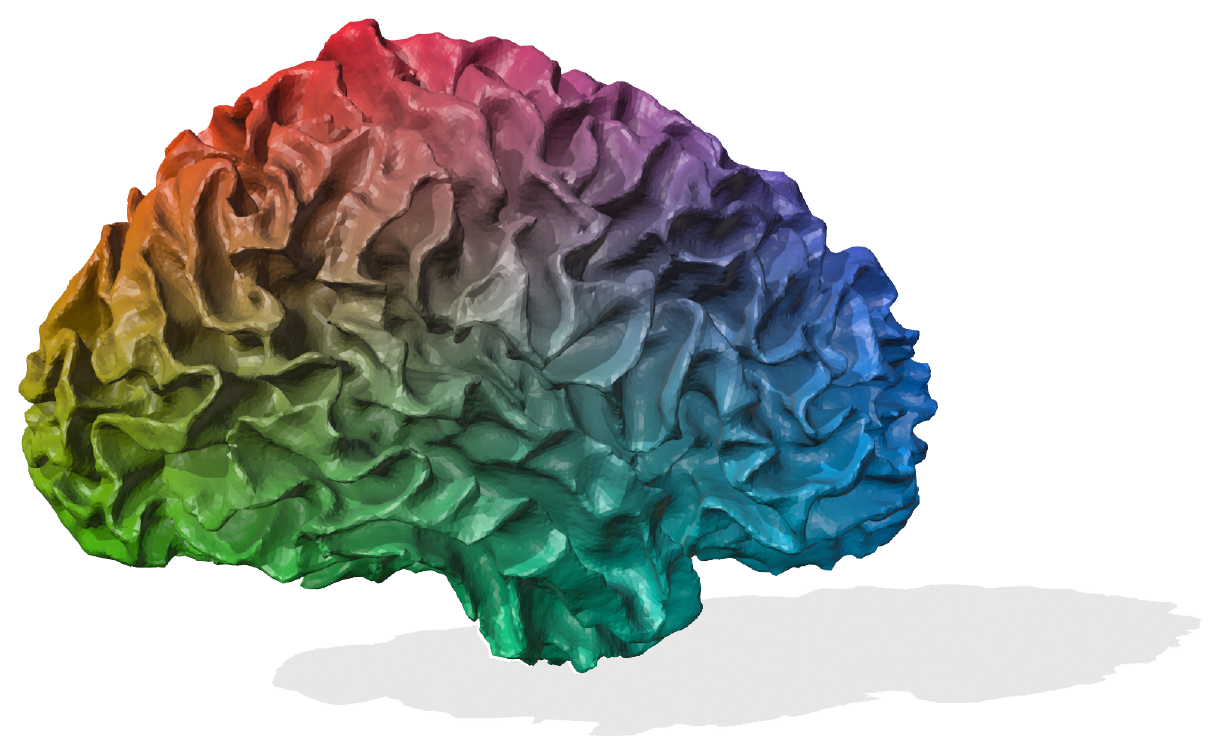}
        \caption{Source}
    \end{subfigure}%
    \hfill
    \begin{subfigure}[t]{0.30\linewidth}
        \includegraphics[width=\linewidth]{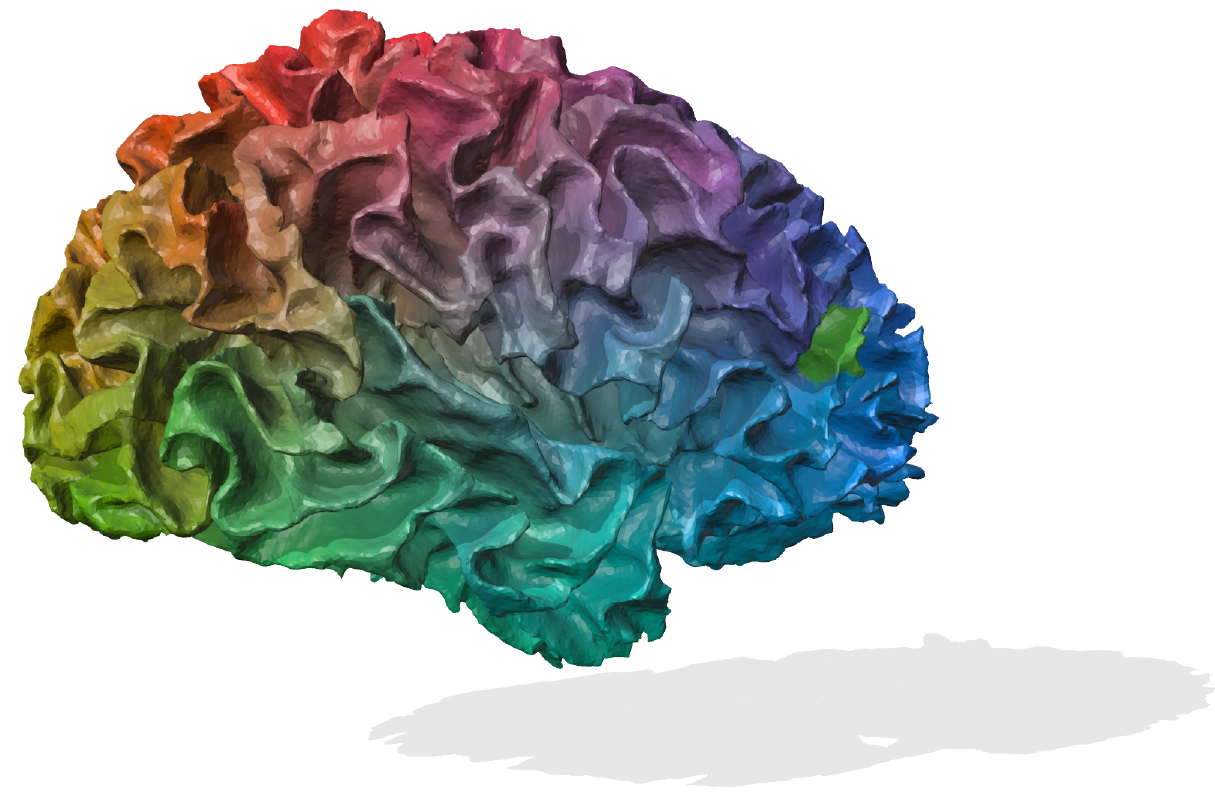}
        \caption{FBM}
    \end{subfigure}%
    \hfill
    \begin{subfigure}[t]{0.30\linewidth}
        \includegraphics[width=\linewidth]{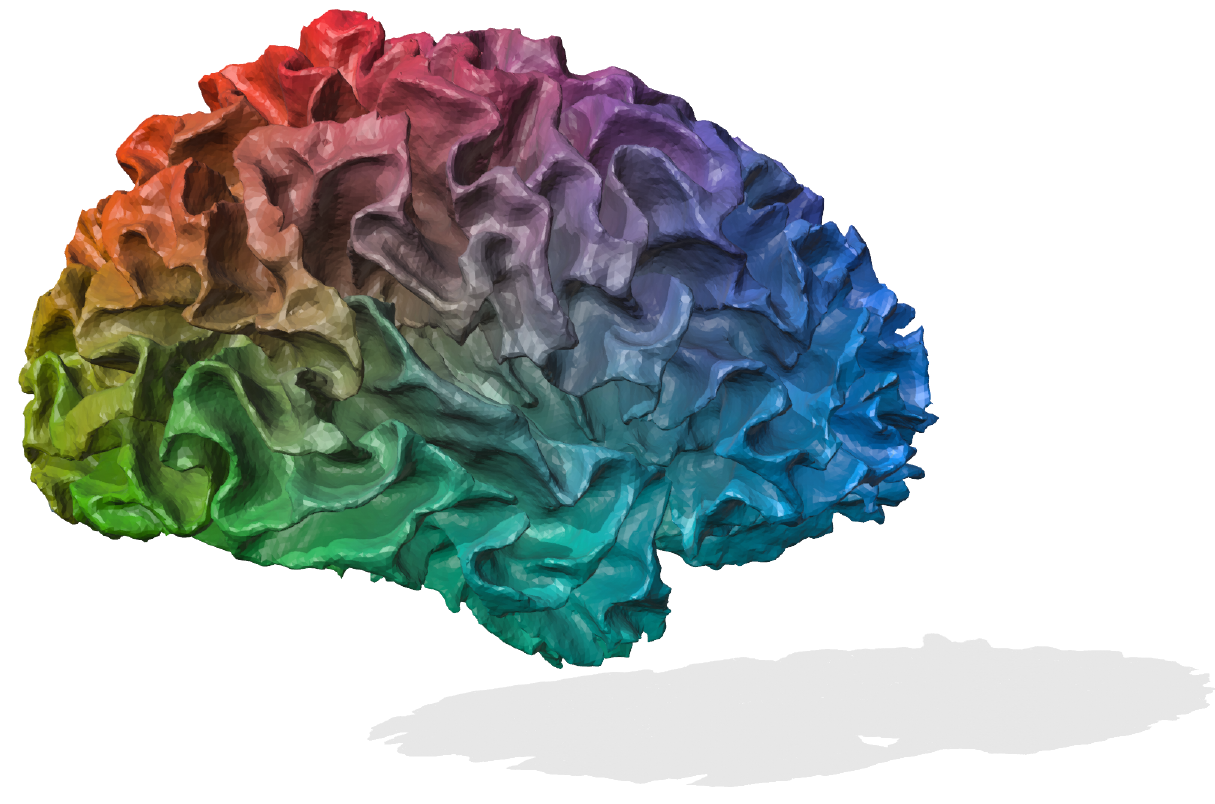}
        \caption{FreeSurfer}
    \end{subfigure}

    \begin{subfigure}[t]{0.30\linewidth}
        \includegraphics[width=\linewidth]{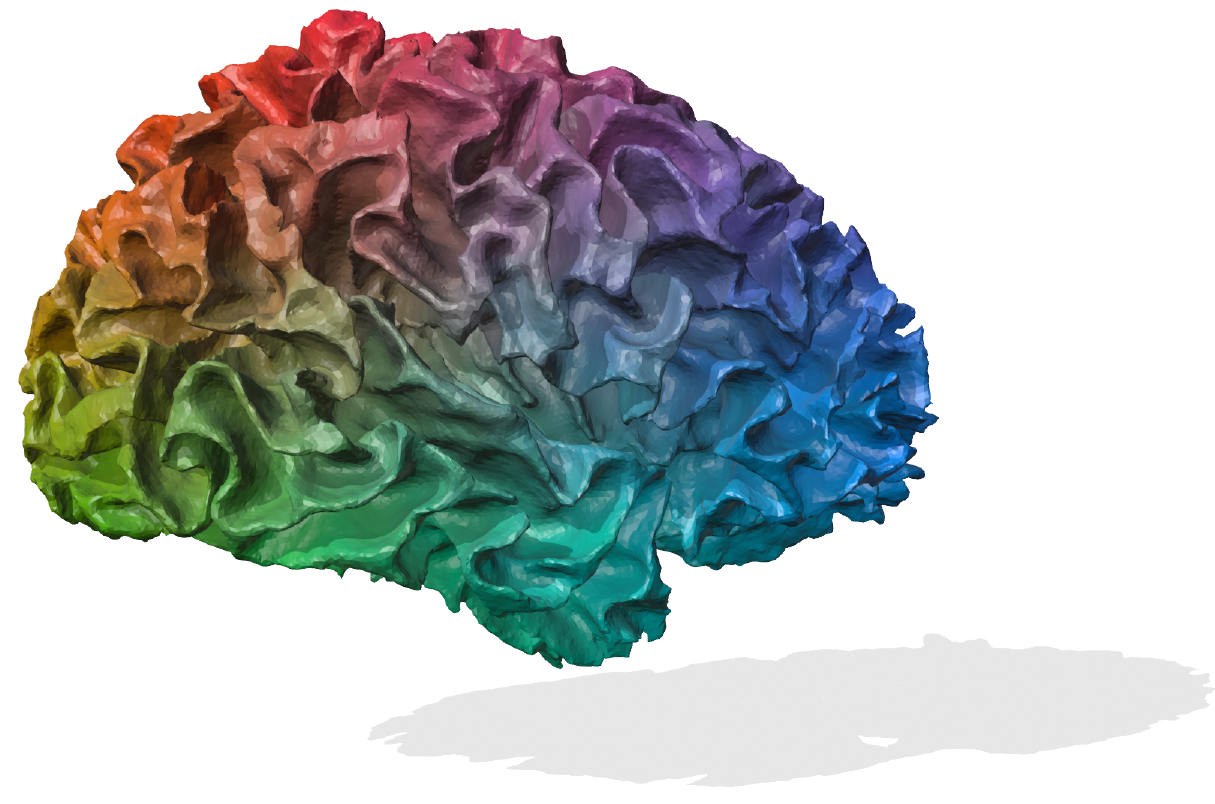}
        \caption{V2C-Flow}
    \end{subfigure}%
    \hspace{0.02\linewidth}
    \begin{subfigure}[t]{0.6\linewidth}
    \centering
        \includegraphics[width=0.54\linewidth]{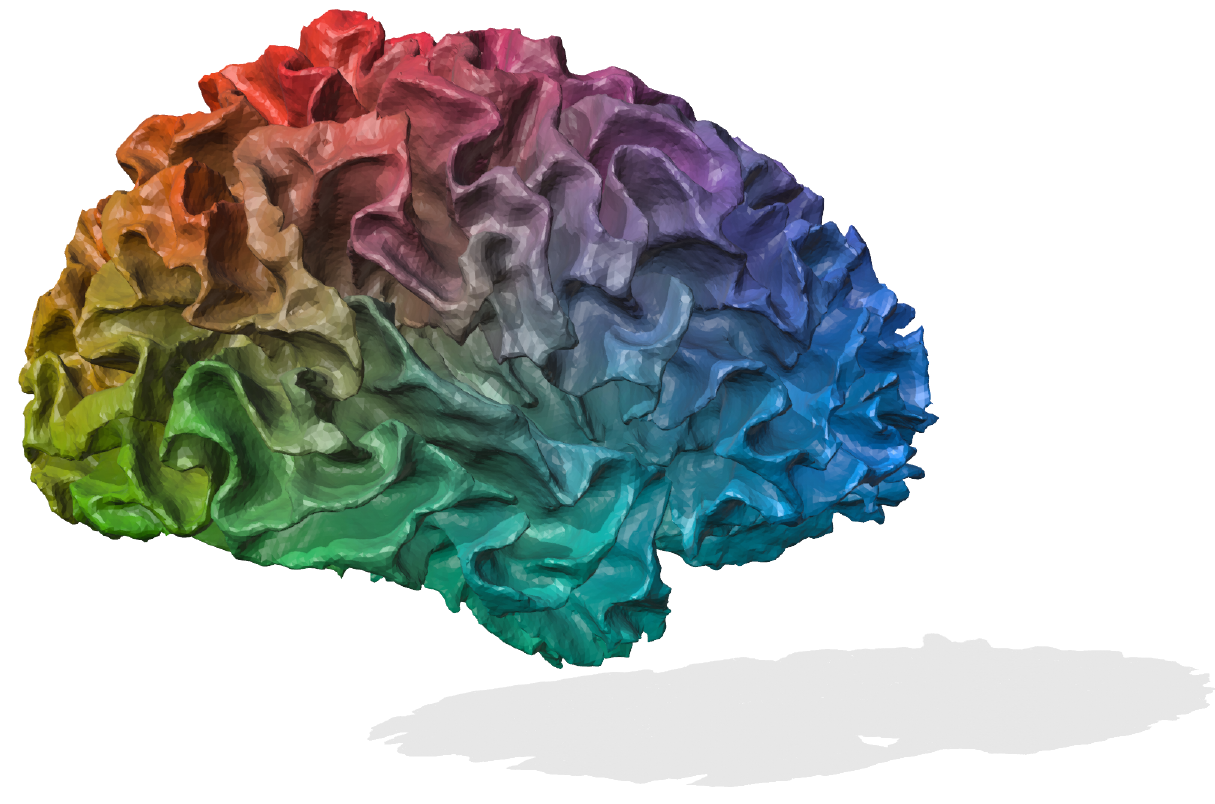}
        \caption{ULRSSM + FlashAtt. (ours)}
    \end{subfigure}

    \caption{\textbf{Smooth-field transfer on Mindboggle-101.} A source field is transferred to a held-out target using each correspondence method.}
    \label{fig:cortex_gradient}
    \vspace{-4mm}
\end{figure}

\begin{figure}[!htb]
    \centering
    \begin{subfigure}[t]{0.30\linewidth}
        \includegraphics[width=\linewidth]{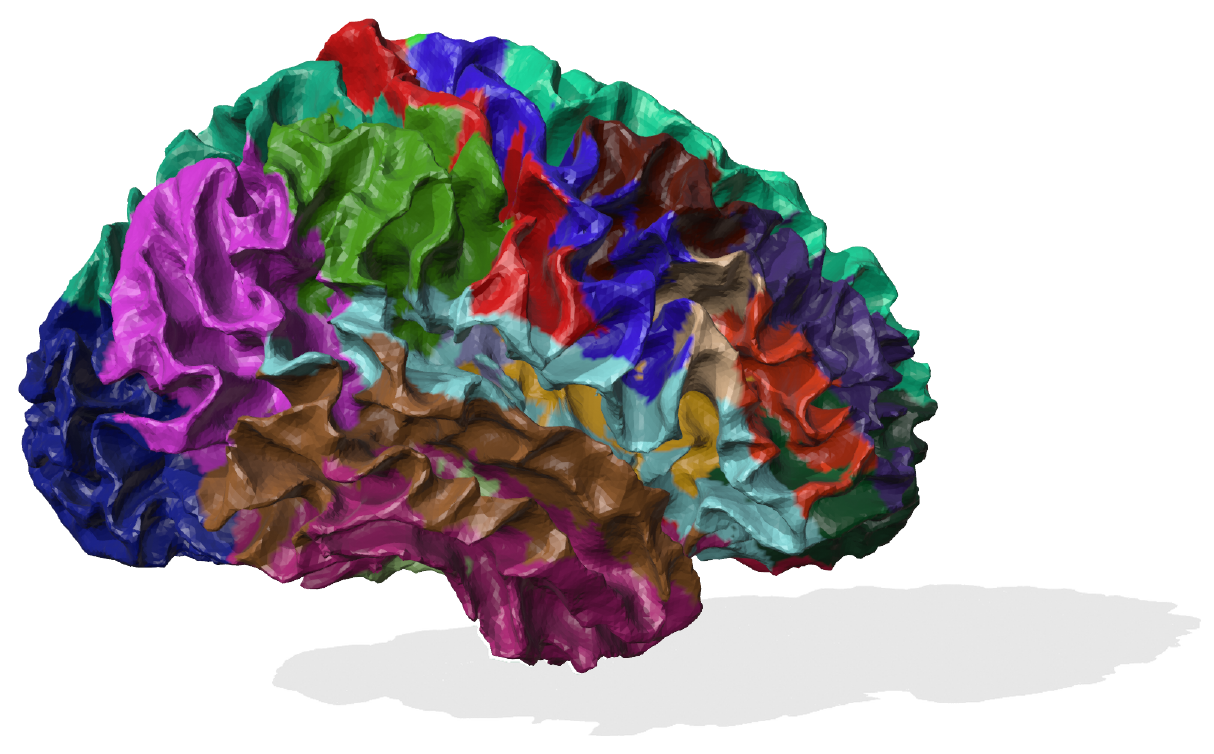}
        \caption{Source}
    \end{subfigure}%
    \hfill
    \begin{subfigure}[t]{0.30\linewidth}
        \includegraphics[width=\linewidth]{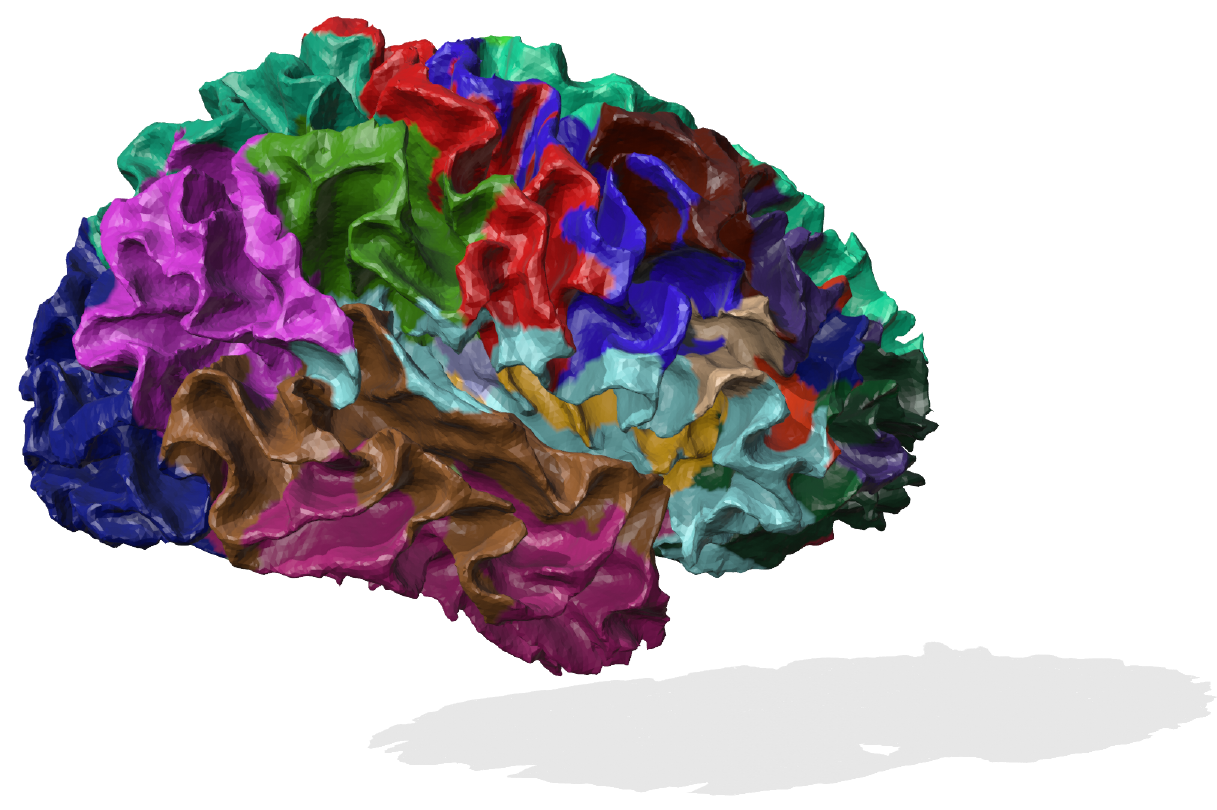}
        \caption{FBM}
    \end{subfigure}%
    \hfill
    \begin{subfigure}[t]{0.30\linewidth}
        \includegraphics[width=\linewidth]{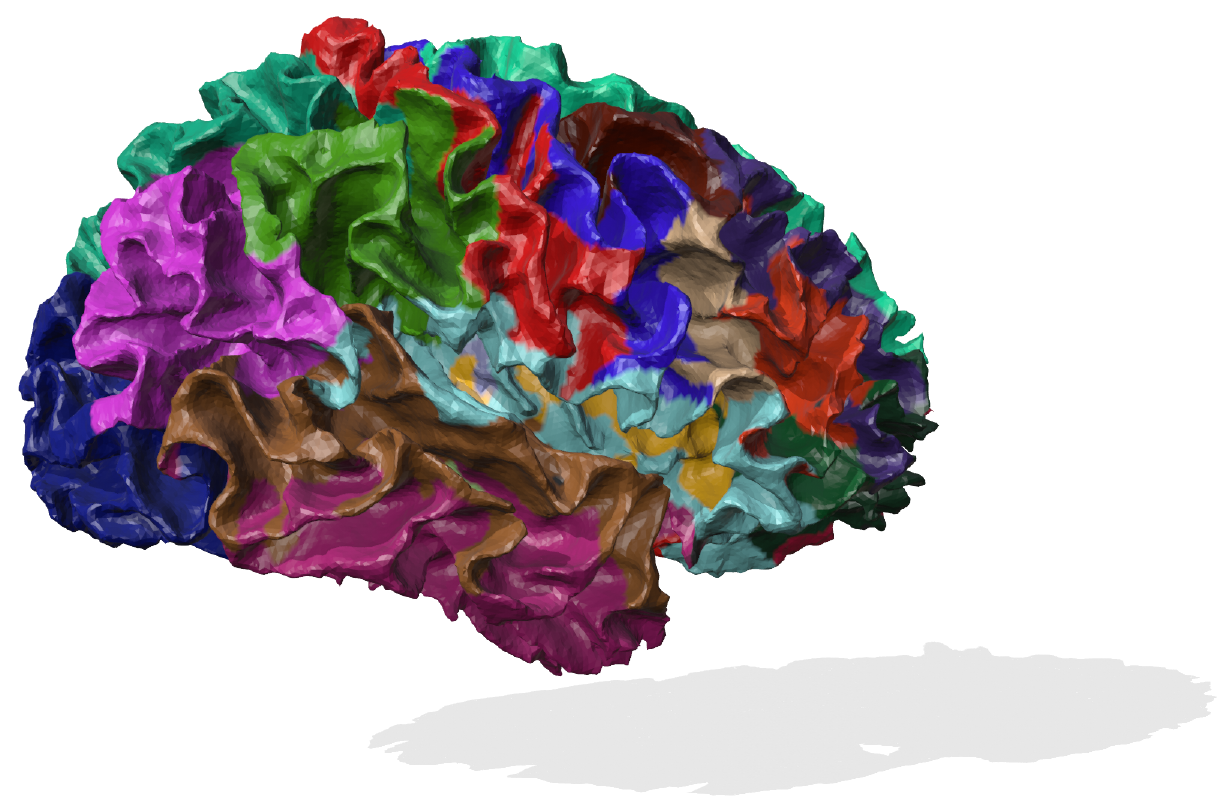}
        \caption{FreeSurfer}
    \end{subfigure}
    \begin{subfigure}[t]{0.30\linewidth}
        \includegraphics[width=\linewidth]{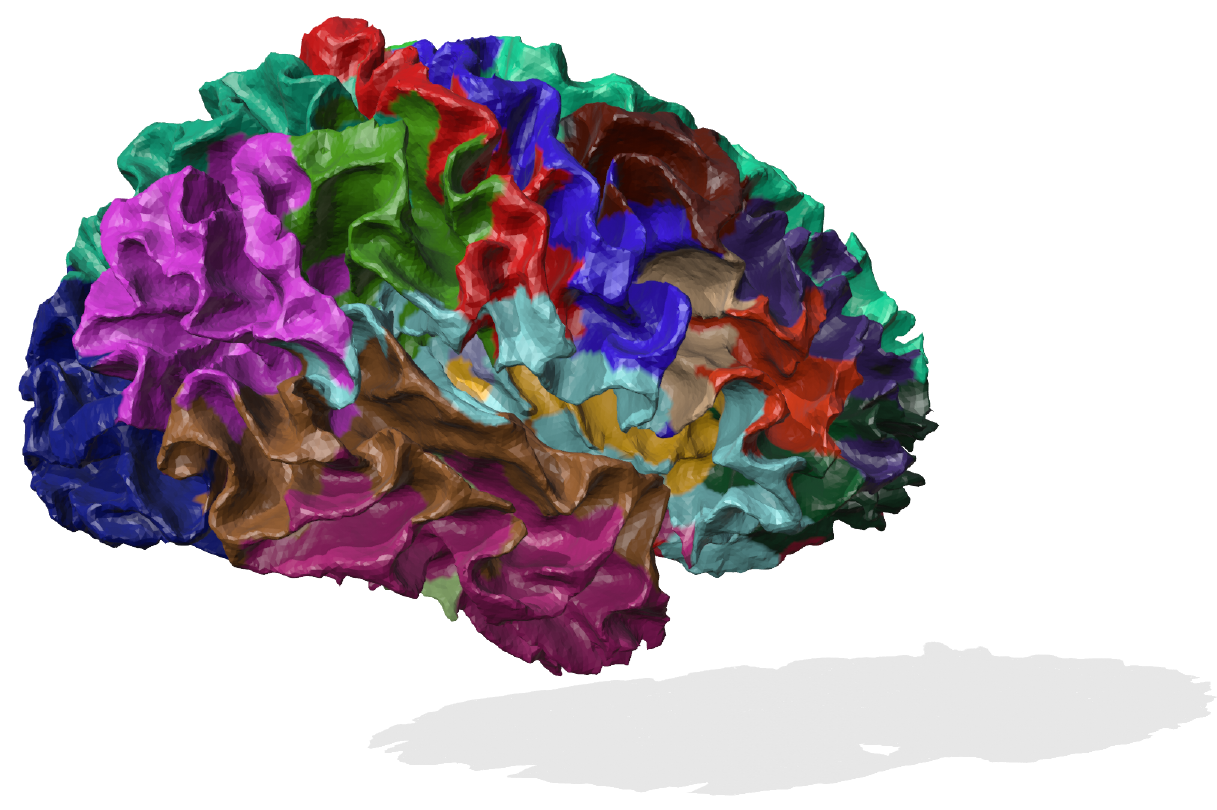}
        \caption{V2C-Flow}
    \end{subfigure}%
    \hfill
    \begin{subfigure}[t]{0.30\linewidth}
        \includegraphics[width=\linewidth]{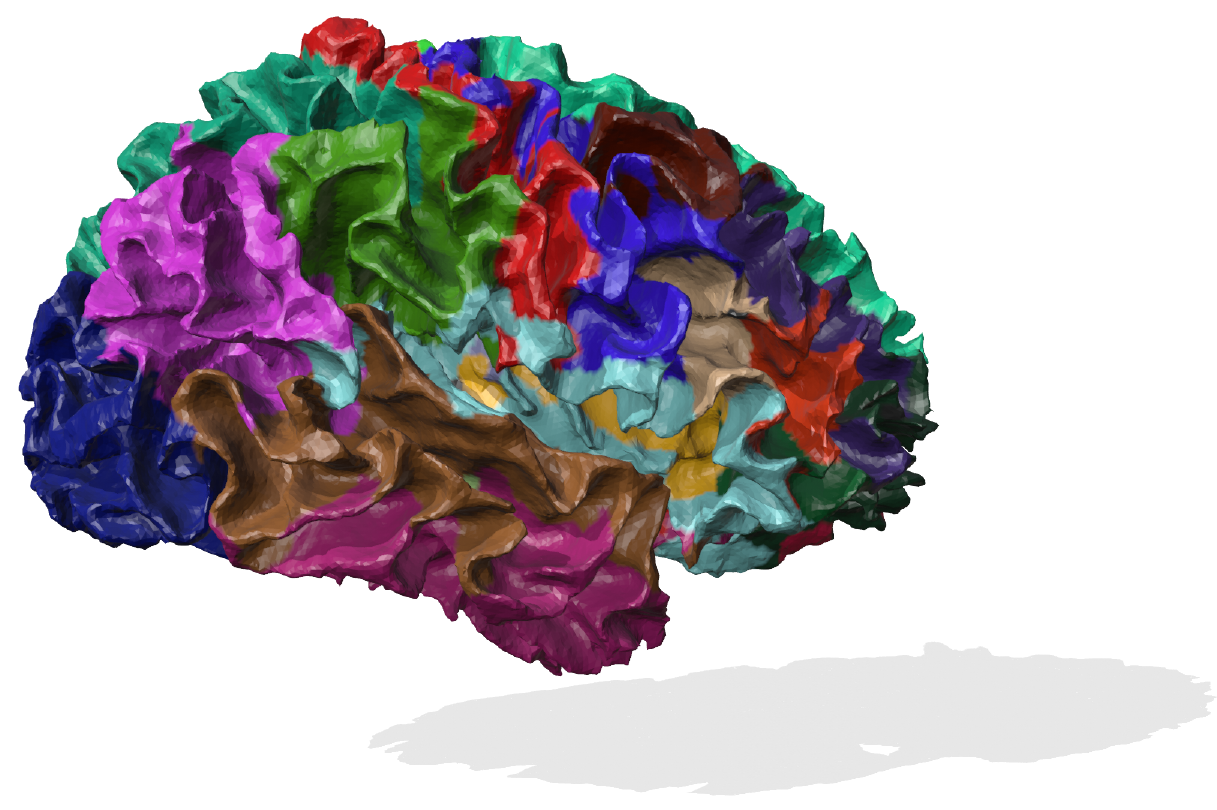}
        \caption{ULRSSM + FlashAtt. (ours)}
    \end{subfigure}%
    \hfill
    \begin{subfigure}[t]{0.30\linewidth}
        \includegraphics[width=\linewidth]{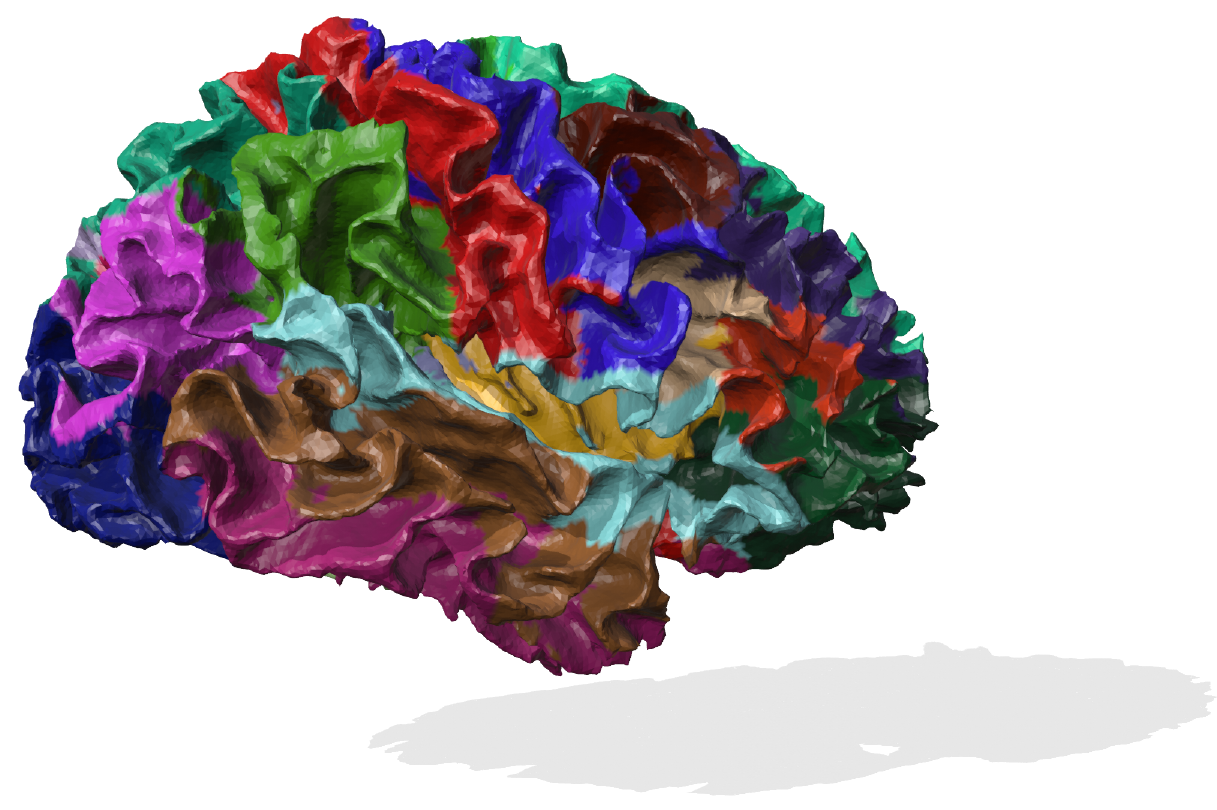}
        \caption{Ground truth}
    \end{subfigure}
    \caption{\textbf{Cortical parcellation transfer on Mindboggle-101.} DKT31 source labels are transferred to the same target; the final panel shows its ground-truth atlas.}
    \label{fig:cortex_atlas}
    \vspace{-4mm}
\end{figure}

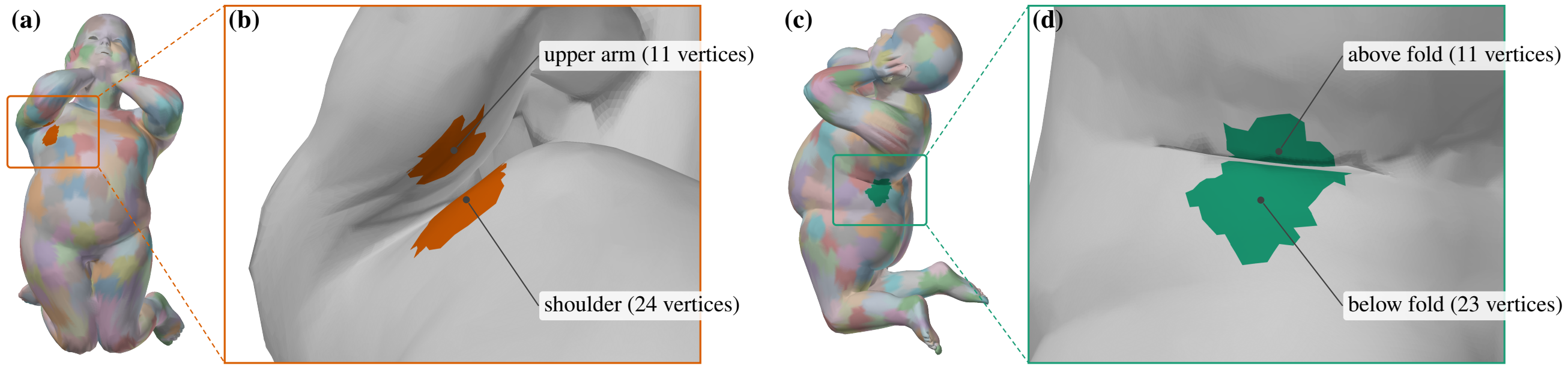
\begin{figure*}[!t]
    \centering
    \resizebox{\textwidth}{!}{\input{img/qualitative/topkids_euclidean_patches.tex}}
    \caption{\textbf{Euclidean patches on self-contact regions in TOPKIDS.} Euclidean patch partition of a TOPKIDS
    shape (kid06, $353$ patches). (a, c) One patch highlighted in each; (b, d) close-ups of the
    boxed regions. Neither patch is a connected region: each covers two surface pieces that
    touch in $\mathbb{R}^3$ but are separated along the surface, the underside of the upper arm
    and the shoulder (b), and the two sides of an abdominal skin fold (d).}
    \label{fig:topkids_patches}
\end{figure*}

\begin{table*}[!t]
\centering
\begin{minipage}[t]{0.33\textwidth}
\centering
\captionof{table}{\textbf{Euclidean vs.\ geodesic patches.} P1: vertices whose assigned
patch is not their geodesically nearest. P2: patches that are not connected regions, and
vertices outside the largest piece of their patch. Averages over all meshes of each
benchmark; the geodesic partition has none by construction. Violations are common and do
not grow with resolution.}
\label{tab:patch_geometry}
\footnotesize
\setlength{\tabcolsep}{3pt}
\resizebox{\linewidth}{!}{%
\begin{tabular}{lrrrr}
\toprule
& & P1 viol. & \multicolumn{2}{c}{P2 viol.\ (\%)} \\
\cmidrule(lr){4-5}
Benchmark & $|V|$ & (\%) & patches & vertices \\
\midrule
FAUST      & 5.0\,k   & 10.6 & \phantom{0}9.2 & 4.4 \\
SCAPE      & 5.0\,k   & \phantom{0}6.1 & \phantom{0}2.2 & 0.6 \\
SHREC'19   & 5.2\,k   & 10.1 & \phantom{0}8.3 & 4.0 \\
SMAL       & 5.2\,k   & 13.0 & 13.3 & 5.8 \\
TOPKIDS    & 10.4\,k  & 12.3 & 11.2 & 2.6 \\
TOPKIDS-HD & 60\,k    & 11.3 & \phantom{0}6.2 & 1.6 \\
Cortex     & 146\,k   & \phantom{0}8.0 & 14.8 & 2.9 \\
\bottomrule
\end{tabular}}
\end{minipage}\hfill
\begin{minipage}[t]{0.27\textwidth}
\centering
\captionof{table}{\textbf{Support quality under either partition} on SMAL ($400$ test
pairs), by candidate budget $R$. Recall: ground-truth correspondent inside the support.
Mass: dense softmax mass the support retains. The two partitions differ marginally.}
\label{tab:patch_recall}
\footnotesize
\setlength{\tabcolsep}{3pt}
\resizebox{\linewidth}{!}{%
\begin{tabular}{rrrrrr}
\toprule
& & \multicolumn{2}{c}{Recall} & \multicolumn{2}{c}{Mass} \\
\cmidrule(lr){3-4}\cmidrule(lr){5-6}
$R$ & $|\mathcal S_{a,R}|$ & Euc. & Geo. & Euc. & Geo. \\
\midrule
\phantom{0}4 & \phantom{0}132 & 0.685 & 0.698 & 0.716 & 0.735 \\
\phantom{0}8 & \phantom{0}260 & 0.829 & 0.841 & 0.860 & 0.879 \\
12 & \phantom{0}379 & 0.882 & 0.888 & 0.907 & 0.921 \\
16 & \phantom{0}487 & 0.909 & 0.913 & 0.931 & 0.941 \\
24 & \phantom{0}692 & 0.939 & 0.940 & 0.954 & 0.961 \\
40 & 1078 & 0.964 & 0.965 & 0.973 & 0.977 \\
64 & 1632 & 0.979 & 0.980 & 0.984 & 0.987 \\
\bottomrule
\end{tabular}}
\end{minipage}\hfill
\begin{minipage}[t]{0.34\textwidth}
\centering
\captionof{table}{\textbf{Restricted vs.\ dense coupling.} Top:
$\|\hat C - C\|_F/\|C\|_F$ between the functional maps computed through the restricted
and the dense coupling. Bottom: dense argmax retained at $R=40$, the criterion budget $R^\ast$,
and point-map agreement with dense at $R=40$. The error decreases smoothly with $R$
and is largest on the non-isometric benchmarks.}
\label{tab:R_ladder}
\footnotesize
\setlength{\tabcolsep}{3pt}
\resizebox{\linewidth}{!}{%
\begin{tabular}{rrrrrr}
\toprule
$R$ & FAUST & SCAPE & SHREC'19 & SMAL & DT4D \\
\midrule
\phantom{00}4 & $5.4$e-2 & $4.9$e-2 & $2.1$e-1 & $1.3$e-1 & $4.8$e-2 \\
\phantom{00}8 & $1.4$e-2 & $1.7$e-2 & $1.2$e-1 & $5.7$e-2 & $1.2$e-2 \\
\phantom{0}16 & $6.8$e-3 & $1.1$e-2 & $7.0$e-2 & $2.9$e-2 & $5.5$e-3 \\
\phantom{0}32 & $4.0$e-3 & $6.6$e-3 & $3.9$e-2 & $1.8$e-2 & $3.4$e-3 \\
\phantom{0}40 & $3.3$e-3 & $5.6$e-3 & $3.2$e-2 & $1.4$e-2 & $3.0$e-3 \\
\phantom{0}64 & $1.9$e-3 & $3.6$e-3 & $1.8$e-2 & $6.5$e-3 & $2.0$e-3 \\
128 & $5.6$e-4 & $1.4$e-3 & $5.2$e-3 & $1.5$e-3 & $9.3$e-4 \\
\midrule
\multicolumn{6}{l}{\footnotesize dense argmax retained at $R=40$} \\
& 1.000 & 1.000 & 0.997 & 0.987 & 1.000 \\
\multicolumn{6}{l}{\footnotesize $R^\ast$, smallest budget retaining $99\%$} \\
& \phantom{0}8 & \phantom{0}8 & 24 & 56 & \phantom{0}8 \\
\multicolumn{6}{l}{\footnotesize map agreement with dense at $R=40$} \\
& 0.992 & 0.995 & 0.949 & 0.960 & 0.994 \\
\bottomrule
\end{tabular}}
\end{minipage}
\end{table*}

\begin{figure*}[!t]
\centering
\includegraphics[width=0.85\textwidth]{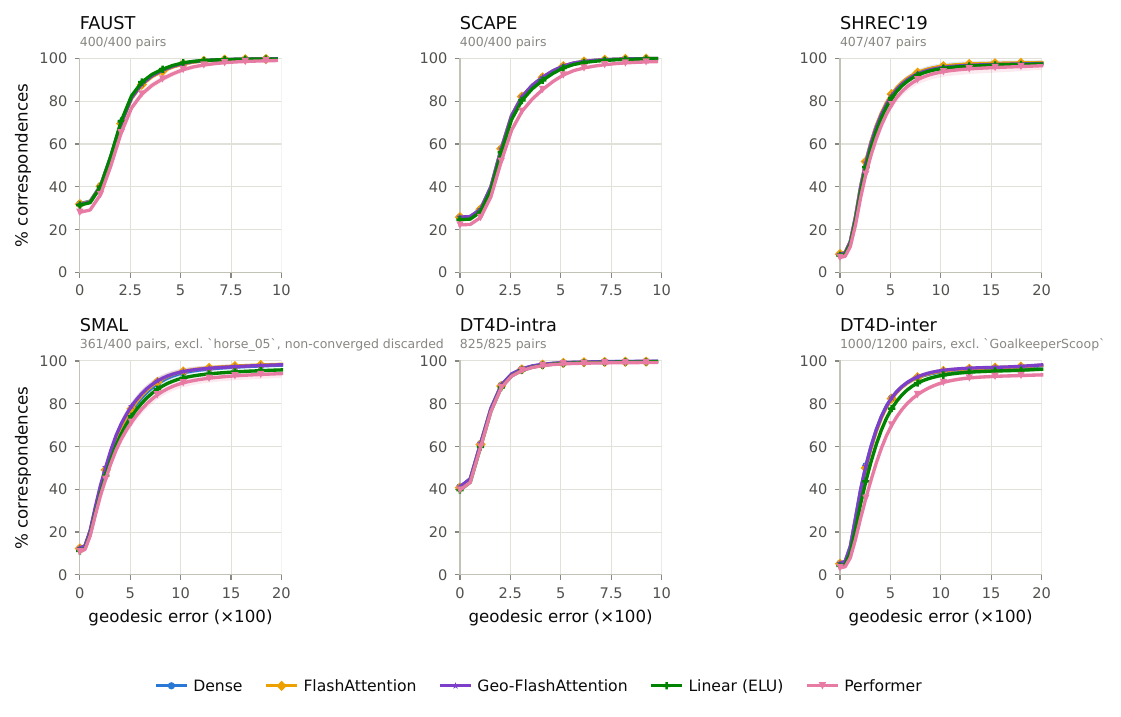}
\caption{\textbf{Geodesic error distributions.} Cumulative fraction of correspondences
below each error threshold, using the same runs and exclusions as
\main{tab:main,tab:linear}. Near-isometric panels span $0$--$10$ and the others
$0$--$20$; pair counts and exclusions are shown in each panel.}
\label{fig:pck_curves}
\end{figure*}

\subsection{Coupling Sharpness and Non-Isometry}
\label{app:sharpness}

The temperature $\tau{=}0.07$ of the soft correspondence (\main{sec:method}) is inherited from prior work \citep{cao2023unsupervised} and held fixed in every experiment.
Here, we ask whether that constant is optimal, whether its optimum is constant, and how this sharpness affects linear couplings.

To vary sharpness continuously around a specific temperature, we replace the exponential kernel on the normalized descriptors with the bandwidth-matched spherical kernel $((1+q_i^\top k_j)/2)^p$, which approximates $\exp(q_i^\top k_j/\tau)$ on the unit sphere and matches $\tau$ at $p = 2/\tau \approx 29$.
A larger $p$ is sharper.
We assess the product using a single converged run per benchmark and a fixed subset of test pairs, measuring the alignment loss of the induced functional map relative to the exact softmax.

\paragraph{Results}
We observe that the optimal sharpness depends on the benchmark and even changes sign with non-isometry. 
On FAUST the objective falls to $0.67$ of the softmax value at $p \approx 96$; SCAPE, DT4D and TOPKIDS-HD show optima near $p \approx 48$ ($0.95$--$0.96$). 
On the strongly non-isometric SMAL, sharpening beyond the matched point only hurts ($1.06$ at $p{=}32$, $1.39$ at $p{=}128$).
These results suggest that near-isometric matching prefers an almost unimodal coupling, while non-isometric matching relies on the blur of the softer softmax.

We also learn $p$ jointly with the descriptors, selecting checkpoints with the
sharpness-independent hard assignment. On SMAL, all five seeds converge to
$p\approx63$ (within $\pm1$) and reduce validation error relative to the fixed-$\tau$
softmax in every seed, by $9\,\%$ on average.
On FAUST, $p$ continues to increase under the shorter schedule but gives slightly worse
validation error than the fixed-$\tau$ softmax. Thus the preferred sharpness is
benchmark-dependent and changes as the descriptors co-adapt. 
These diagnostic results are not directly comparable to \main{tab:main}, where we keep $\tau$ fixed in the reported comparisons because optimizing sharpness modulates the coupling rather than its implementation.

%% file: img/qualitative/topkids_euclidean_patches.tex
\definecolor{contactArm}{HTML}{D95F02}
\definecolor{contactFold}{HTML}{1B9E77}
\begin{tikzpicture}[
    panel/.style={inner sep=0pt, outer sep=0pt, anchor=south west},
    letter/.style={anchor=north west, inner sep=1pt, font=\footnotesize\bfseries},
    region/.style={draw=#1, line width=0.6pt, rounded corners=1pt},
    zoomline/.style={draw=#1, line width=0.4pt, dash pattern=on 1.5pt off 1pt},
    frame/.style={draw=#1, line width=0.6pt},
    callout/.style={font=\scriptsize, text=black, fill=white, fill opacity=0.85,
                    text opacity=1, rounded corners=1pt, inner sep=1.5pt},
    leader/.style={draw=black!75, line width=0.4pt},
    piece/.style={circle, fill=black!75, inner sep=0pt, minimum size=2.2pt},
]
\def\panelH{3.9cm}
\def\gap{3.5mm}
\newcommand{\panelscope}[1]{%
    \begin{scope}[shift={(#1.south west)},
                  x={($(#1.south east)-(#1.south west)$)},
                  y={($(#1.north west)-(#1.south west)$)}]}

\node[panel] (a) at (0,0)
    {\includegraphics[height=\panelH]{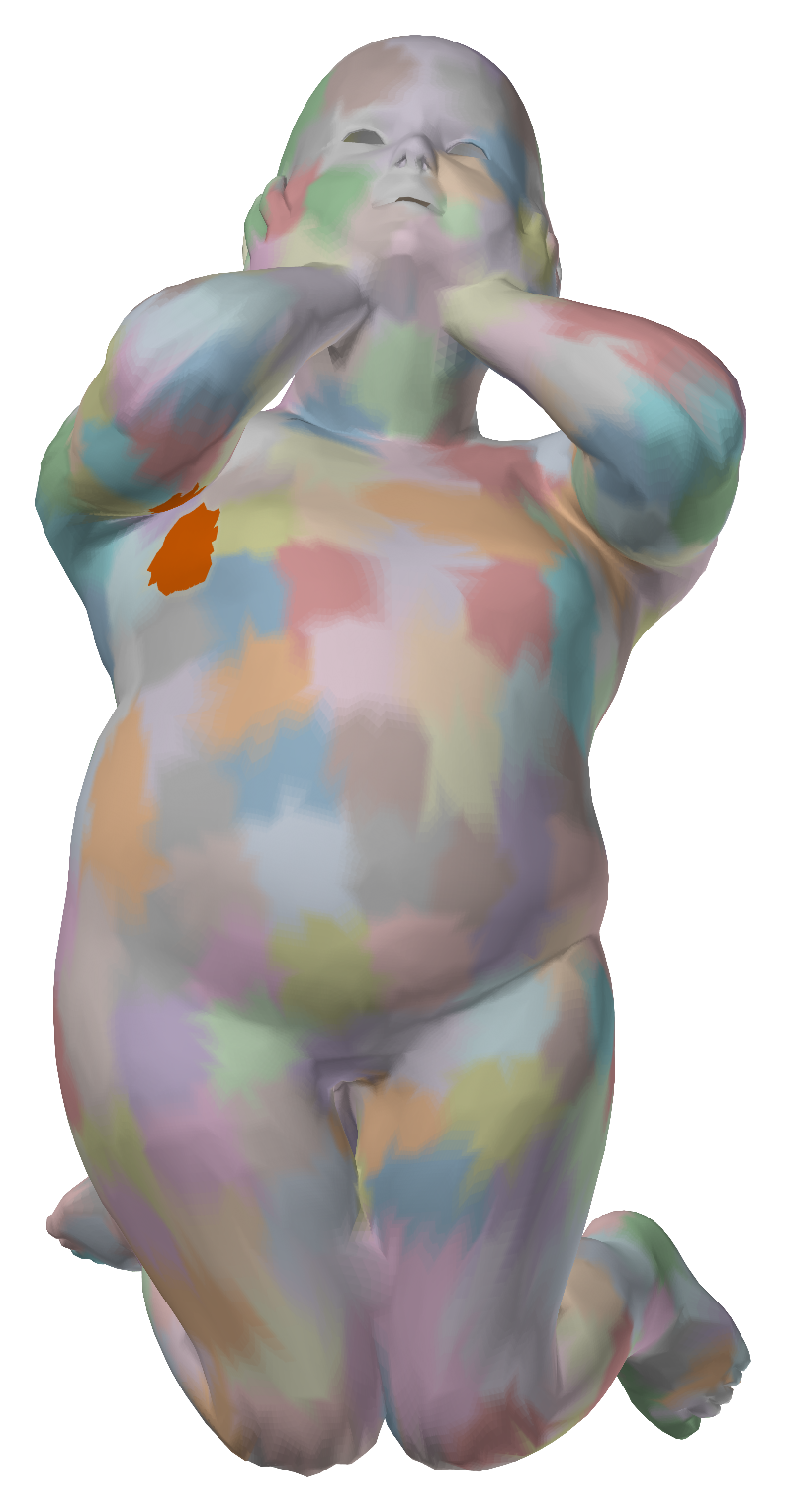}};
\node[panel] (b) at ([xshift=\gap]a.south east)
    {\includegraphics[height=\panelH]{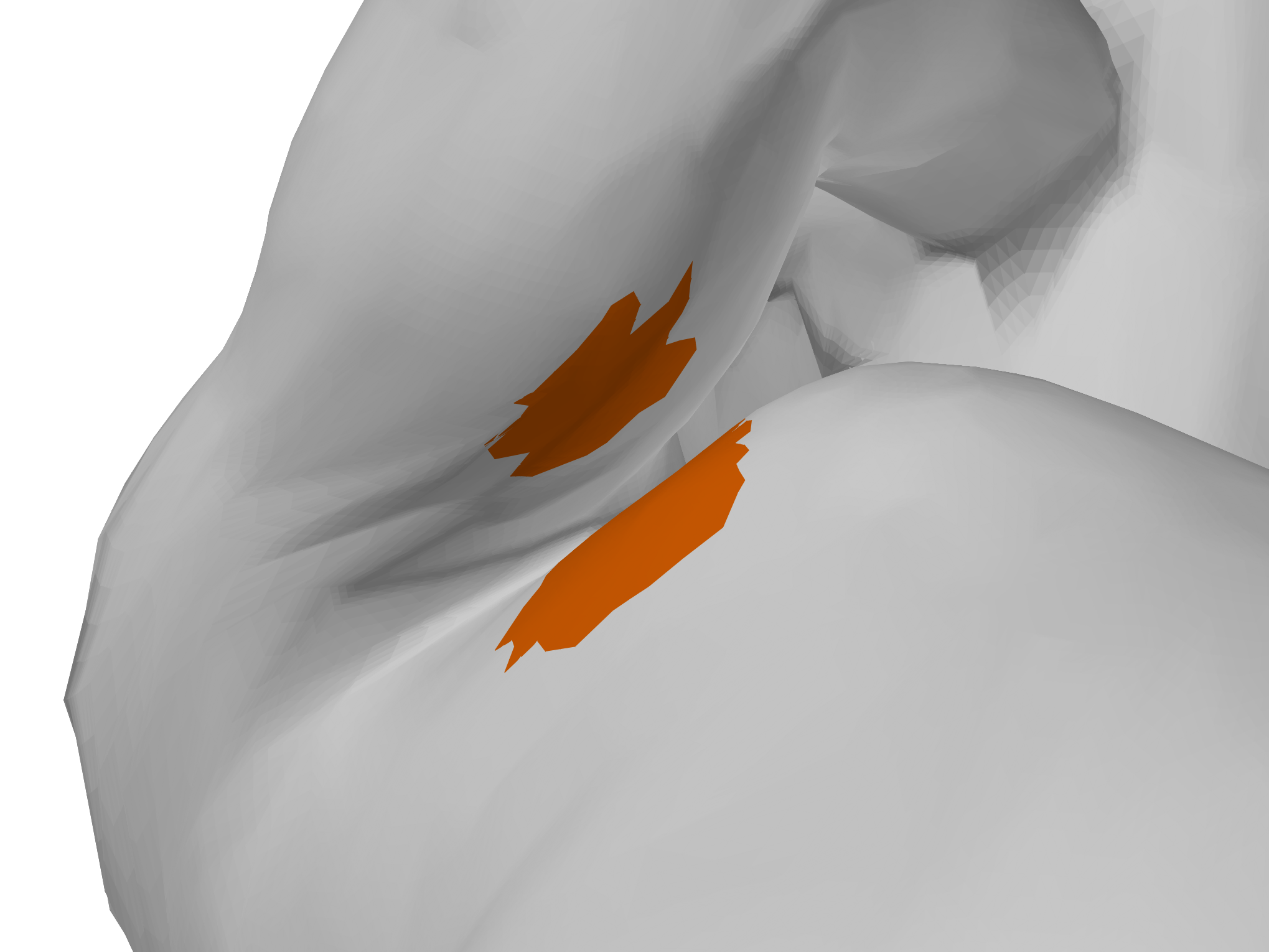}};
\node[panel] (c) at ([xshift=2.5*\gap]b.south east)
    {\includegraphics[height=\panelH]{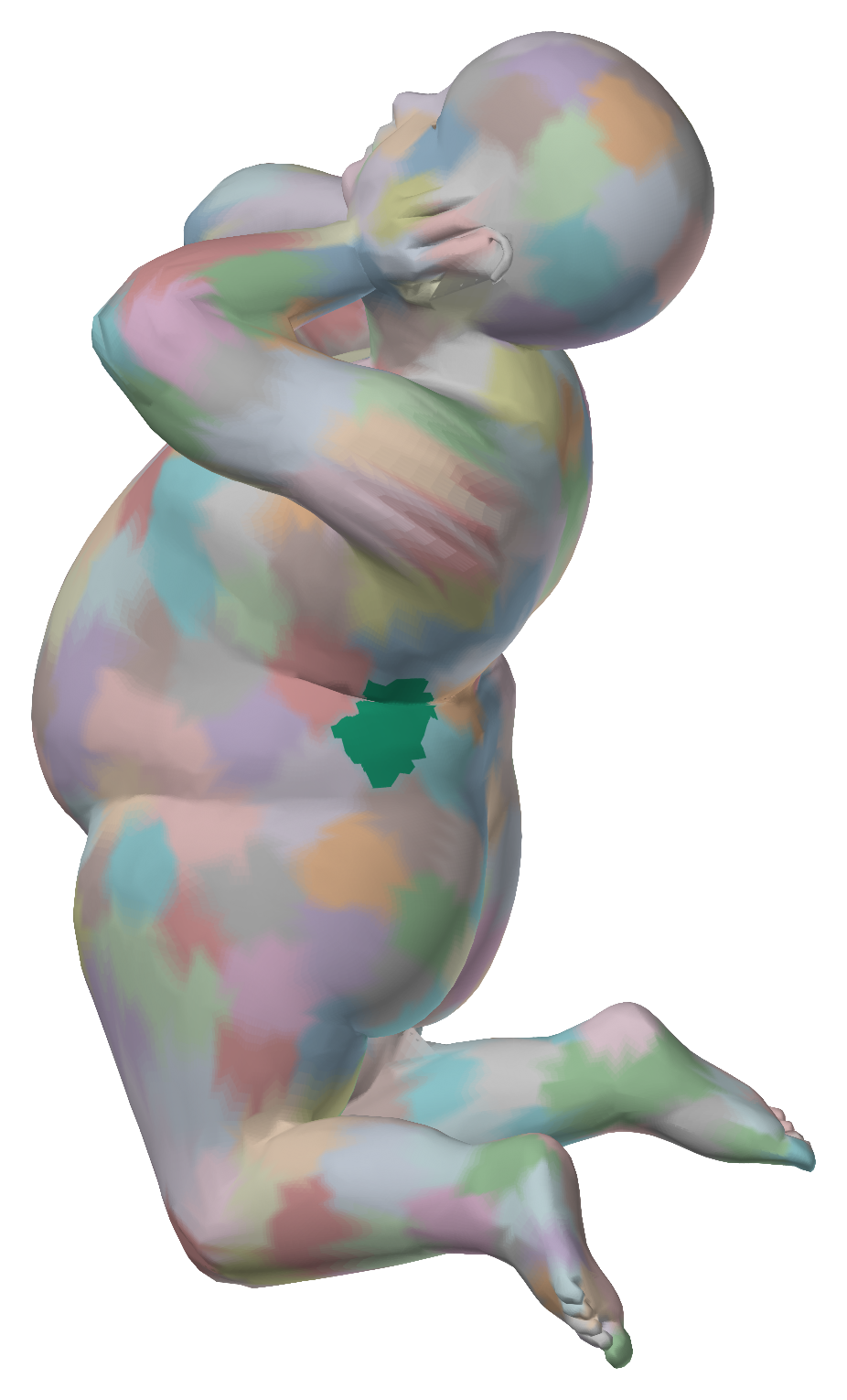}};
\node[panel] (d) at ([xshift=\gap]c.south east)
    {\includegraphics[height=\panelH]{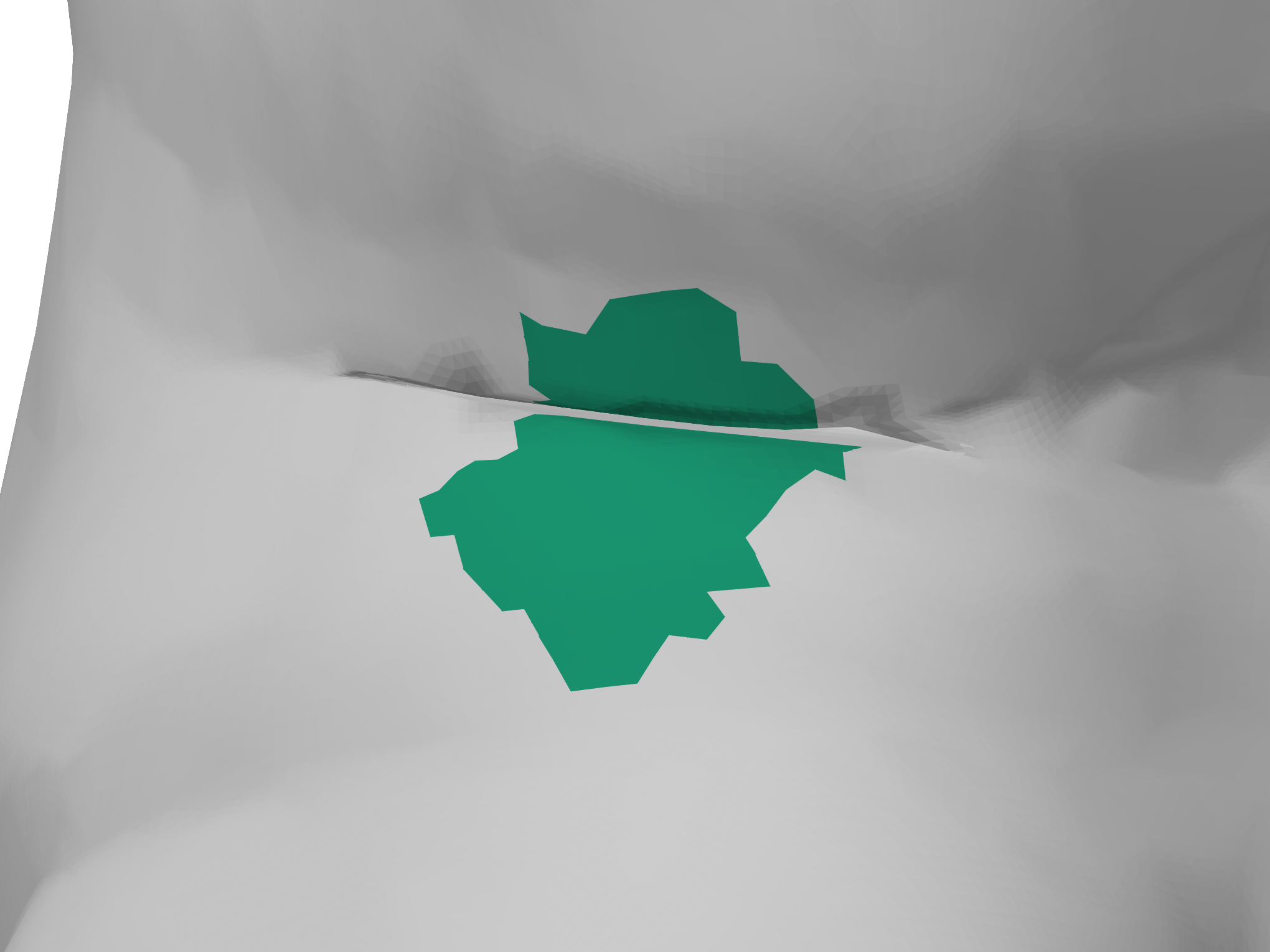}};

\panelscope{a}
    \draw[region=contactArm] (0.000,0.547) rectangle (0.493,0.747);
    \coordinate (aNE) at (0.493,0.747);
    \coordinate (aSE) at (0.493,0.547);
\end{scope}
\draw[frame=contactArm] (b.south west) rectangle (b.north east);
\draw[zoomline=contactArm] (aNE) -- (b.north west);
\draw[zoomline=contactArm] (aSE) -- (b.south west);
\panelscope{b}
    \node[piece] (b1) at (0.482,0.595) {};
    \node[piece] (b2) at (0.508,0.457) {};
    \node[callout, anchor=west] (b1l) at (0.66,0.86) {upper arm (11 vertices)};
    \node[callout, anchor=west] (b2l) at (0.66,0.16) {shoulder (24 vertices)};
\end{scope}
\draw[leader] (b1) -- (b1l.west);
\draw[leader] (b2) -- (b2l.west);

\panelscope{c}
    \draw[region=contactFold] (0.244,0.386) rectangle (0.676,0.582);
    \coordinate (cNE) at (0.676,0.582);
    \coordinate (cSE) at (0.676,0.386);
\end{scope}
\draw[frame=contactFold] (d.south west) rectangle (d.north east);
\draw[zoomline=contactFold] (cNE) -- (d.north west);
\draw[zoomline=contactFold] (cSE) -- (d.south west);
\panelscope{d}
    \node[piece] (d1) at (0.526,0.592) {};
    \node[piece] (d2) at (0.488,0.456) {};
    \node[callout, anchor=west] (d1l) at (0.66,0.86) {above fold (11 vertices)};
    \node[callout, anchor=west] (d2l) at (0.66,0.16) {below fold (23 vertices)};
\end{scope}
\draw[leader] (d1) -- (d1l.west);
\draw[leader] (d2) -- (d2l.west);

\foreach \p in {a,b,c,d} \node[letter] at (\p.north west) {(\p)};
\end{tikzpicture}

%% file: sec/A4_qualitative.tex
\subsection{Qualitative Results}
\label{app:qualitative}

\paragraph{Cortical surfaces}
\Cref{fig:cortex_gradient,fig:cortex_atlas} provide complementary views of a held-out
Mindboggle-101 pair: the smooth field reveals local distortion, while the DKT31 labels
show region-level alignment. ULRSSM with FlashAttention produces coherent fields and
region boundaries comparable to the specialized baselines, consistent with the
quantitative results in \main{sec:cortex}.

%% file: sec/A5_patches.tex
\subsection{Patch Construction: Euclidean and Geodesic Voronoi Cells}
\label{app:patch_metric}

We compare Euclidean patches built from vertex positions with the intrinsic \emph{geodesic}
Voronoi partition used in \main{sec:geo-flash}, in which every vertex belongs to the center nearest along the
surface (P1) and every cell is a connected region (P2). \Cref{tab:patch_geometry}
measures how often Euclidean patches violate these two properties, and
\cref{tab:patch_recall} what this costs the coupling. Our released implementation
builds geodesic patches by default and retains the Euclidean ones for completeness.

\paragraph{TOPKIDS self-contact}
By grouping spatially adjacent surface parts across extrinsic self-contact, Euclidean patch embeddings may regularize the coupling on TOPKIDS (\cref{fig:topkids_patches}).

\subsection{Candidate Budget \texorpdfstring{$R$}{R}}
\label{app:R_budget}

The budget $R$ of Geo-FlashAttention fixes only the support of the sparse attention.
Sweeping it traces how closely the restricted coupling approximates the dense one.
\Cref{tab:R_ladder} reports, per benchmark, the relative error of the functional map
computed through the restricted coupling against the dense one, the fraction of query
rows whose dense argmax remains inside the support, and the agreement of the recovered
point maps. $R^\ast$ summarizes each sweep as the smallest budget that retains the
dense argmax for $99\%$ of rows.

\subsection{Geodesic Error Distributions}
\label{app:pck}

\Cref{fig:pck_curves} shows that, across benchmarks, the maximum separation among exact
couplings is $0.06$--$2.46$ percentage points over the full error distribution. The
linearized variants lag on SMAL and DT4D-inter but track the exact couplings on FAUST,
SCAPE and DT4D-intra, consistent with \cref{tab:linear}.